\documentclass[10pt,journal,compsoc]{IEEEtran}
\usepackage{epsfig}
\usepackage{graphicx}
\usepackage{amsmath}
\usepackage{amssymb}
\usepackage{multirow}
\usepackage{makecell}
\usepackage{tabulary}
\usepackage{stackengine}

\usepackage{url}
\usepackage[pagebackref=true,breaklinks=true,letterpaper=true,colorlinks,bookmarks=false]{hyperref}

\usepackage[ruled, lined, longend, linesnumbered]{algorithm2e}
\usepackage{setspace}

\usepackage{subcaption}
\usepackage{multirow}
\usepackage{hhline}
\usepackage{enumitem}
\usepackage[T1]{fontenc}
\usepackage{caption}

\ifCLASSOPTIONcompsoc
  \usepackage[nocompress]{cite}
\else
  \usepackage{cite}
\fi
\ifCLASSINFOpdf
\else
\fi
\begin{document}
%
\title{Bi-Calibration Networks for Weakly-Supervised \\ Video Representation Learning}
%
%
%
%

\author{Fuchen~Long,
~Ting~Yao,~\IEEEmembership{Senior Member,~IEEE},
~Zhaofan~Qiu,
~Xinmei~Tian,~\IEEEmembership{Member, ~IEEE},
~Jiebo~Luo,~\IEEEmembership{Fellow,~IEEE},
~and~Tao~Mei,~\IEEEmembership{Fellow,~IEEE}
\IEEEcompsocitemizethanks{\IEEEcompsocthanksitem Ting Yao is the corresponding author.}
\IEEEcompsocitemizethanks{\IEEEcompsocthanksitem
Fuchen~Long, Ting~Yao, Zhaofan~Qiu, and Tao~Mei are with JD Explore Academy, Beijing, China. Xinmei~Tian is with University of Science and Technology of China, Hefei, China. Jiebo~Luo is with University of Rochester. (e-mail: longfc.ustc@gmail.com; tingyao.ustc@gmail.com; zhaofanqiu@gmail.com; tmei@jd.com; xinmei@ustc.edu.cn; jluo@cs.rochester.edu).}
\IEEEcompsocitemizethanks{\IEEEcompsocthanksitem This work was supported by the National Key R\&D Program of China under Grant No. 2020AAA0108600.}
}

%
%

\markboth{IEEE TRANSACTIONS ON PATTERN ANALYSIS AND MACHINE INTELLIGENCE}%
{Long \MakeLowercase{\textit{et al.}}: Bi-Calibration Networks for Weakly-Supervised Video Representation Learning}
%



\IEEEtitleabstractindextext{%
\begin{abstract}
The leverage of large volumes of web videos paired with the searched queries or surrounding texts (e.g., title) offers an economic and extensible alternative to supervised video representation learning. Nevertheless, modeling such weakly visual-textual connection is not trivial due to query polysemy (i.e., many possible meanings for a query) and text isomorphism (i.e., same syntactic structure of different text). In this paper, we introduce a new design of mutual calibration between query and text to boost weakly-supervised video representation learning. Specifically, we present Bi-Calibration Networks (BCN) that novelly couples two calibrations to learn the amendment from text to query and vice versa. Technically, BCN executes clustering on all the titles of the videos searched by an identical query and takes the centroid of each cluster as a text prototype. The query vocabulary is built directly on query words. The video-to-text/video-to-query projections over text prototypes/query vocabulary then start the text-to-query or query-to-text calibration to estimate the amendment to query or text. We also devise a selection scheme to balance the two corrections. Two large-scale web video datasets paired with query and title for each video are newly collected for weakly-supervised video representation learning, which are named as YOVO-3M and YOVO-10M, respectively. 
The video features of BCN learnt on 3M web videos obtain superior results under linear model protocol on downstream tasks. More remarkably, BCN trained on the larger set of 10M web videos with further fine-tuning leads to 1.6\%, and 1.8\% gains in top-1 accuracy on Kinetics-400, and Something-Something V2 datasets over the state-of-the-art TDN, and ACTION-Net methods with ImageNet pre-training. Source code and datasets are available at \url{https://github.com/FuchenUSTC/BCN}.
\end{abstract}

}

\maketitle

\IEEEdisplaynontitleabstractindextext

%
\IEEEpeerreviewmaketitle

\IEEEraisesectionheading{\section{Introduction}\label{sec:introduction}}

%
%
%
%

\IEEEPARstart{W}{ith} the rise of deep learning technologies, there has been a steady momentum of breakthroughs on video representation learning \cite{Carreira:CVPR17,Yan:CVPR20,Simonyan:NIPS14,Yang:CVPR20,Liu:V-Swin,C3D:PAMI,T1D:PAMI,Wang:PAMI20,Gaidon:PAMI13}. The achievements rely heavily on the requirement to have large amount of labeled data for fully-supervised training.
In practice, acquiring the annotations of videos is very expensive and time-consuming.
Therefore, recent research \cite{Ghadiyaram:CVPR19,Li:CPD20} study the alternative regime of web data, which is largely available and freely accessible by search engines, for weakly-supervised learning.
These approaches usually treat the weakly visual-textual connection as a reliable signal and directly maximize the similarity between them or alleviate the challenge of noise (e.g., incorrect or irrelevant text) in web data for training, but seldom explore the inherent property of videos or texts.
For example, is it reasonable that the representations of all the searched videos from an identical query cluster around a single prototype? or if the surrounding texts of two videos are close in representation space, should the representations of the two videos also be in proximity?

\begin{figure}[!tb]
	\centering\includegraphics[width=0.49\textwidth]{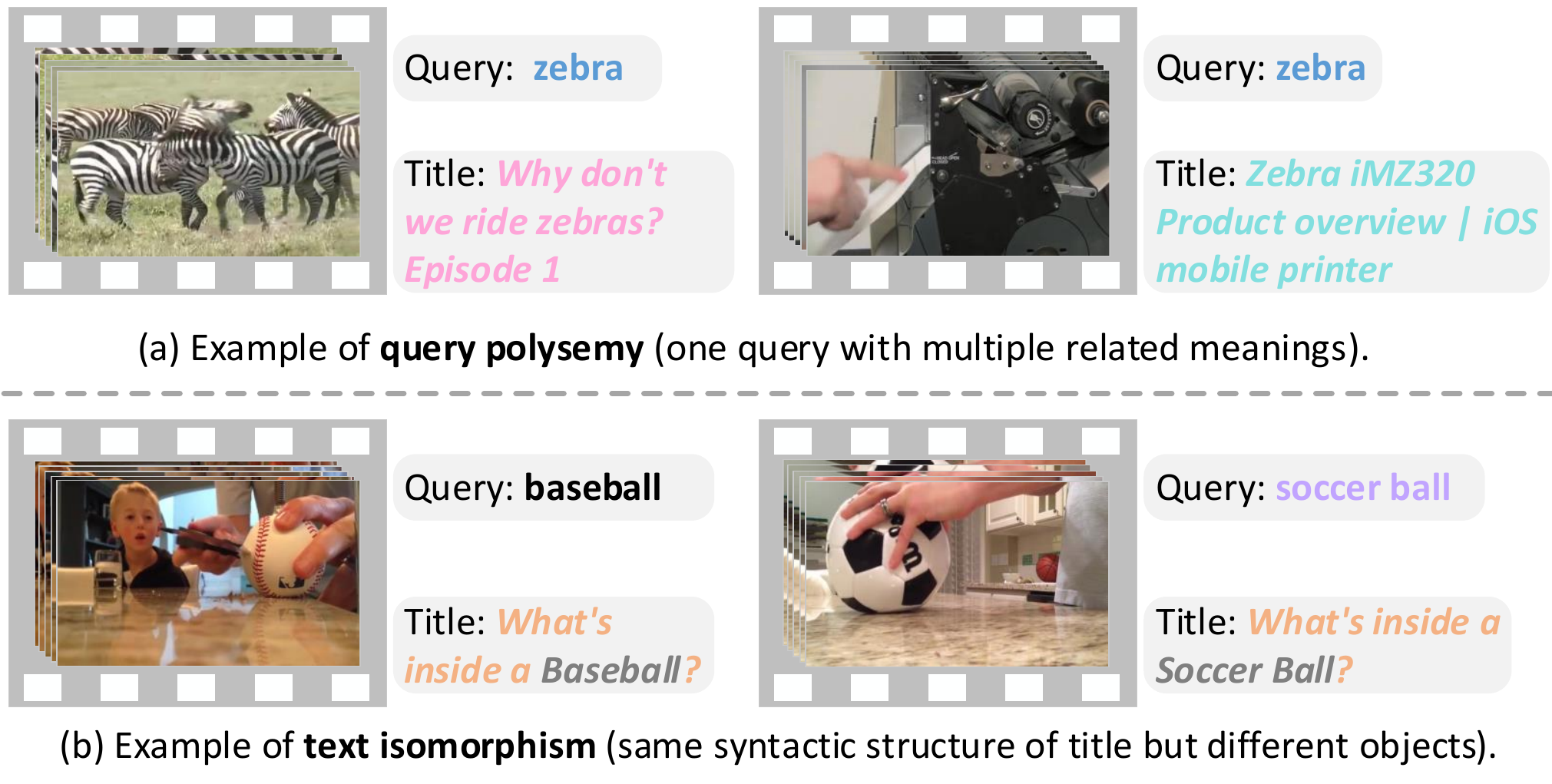}
	\caption{\small The intuitive examples of (a) ``query polysemy'' and (b) ``text isomorphism'' of web videos.}
	\label{fig1:1}
	\vspace{-0.1in}
\end{figure}

In order to answer the two questions, let's look at the two examples illustrated in Figure \ref{fig1:1}. The upper one showcases two returned videos when searching for the query of ``zebra'' in a text-based video search engine. The query ``zebra'' de facto contains two types of search intention: an animal or a brand of printer, and the videos in response to the two meanings are quite distinct in visual appearance. The phenomenon is known as ``query polysemy.'' In this case, taking the two families of videos as one class will inevitably mislead video representation learning through either classification or cross-view embedding. We propose to mitigate the issue by leveraging the calibration from surrounding text (i.e., title in this work) of the videos. The titles of ``why don't we ride zebras? Episode 1'' and ``Zebra iMZ320 Product overview | iOS mobile printer'' provide rich information about the video content and are more descriptive. The videos are naturally grouped into two clusters based on valuable supervision of titles and each cluster corresponds to one semantic meaning, potentially making video representation learning more discriminative. In contrast, robust learning of video representation also necessitates using the query to regulate the text. Taking the lower case as an example, the titles of ``What's inside a Baseball?'' and ``What's inside a Soccer Ball?'' are in the same syntactic structure~but the objectives are different. We define this as ``text isomorphism.'' Solely capitalizing on such titles will tend to draw the two videos close in representation space but unfortunately the videos describe two different objects, which are indicated by search queries. As a result, query information here is a rewarding signal to adjust visual-textual correlation and further improve video representation learning.

By delving into mutual calibration across query and text for weakly-supervised video representation learning, we present a novel Bi-Calibration Networks (BCN) architecture with the backbone of 3D ConvNet.
Specifically, we employ the off-the-shelf BERT \cite{Jacob:ACL18} model to extract the title features and cluster all the titles of the videos searched by an identical query into clusters. The centroid of each cluster is then taken as a text prototype and put into text vocabulary. Primary text supervision is measured as the cosine similarity between video title and all text prototypes in the vocabulary to regulate video-to-text (v2t) projection. Similarly, we take the ``one-hot'' vector in query vocabulary built directly on query words as primary query supervision to optimize video-to-query (v2q) projection. Next, v2t/v2q projection starts the text-to-query (t2q) or query-to-text (q2t) calibration, in which BCN aggregates/decomposes the predictions on cluster/query level in a bottom-up/top-down way to produce the t2q/q2t correction. The two corrections refine the primary query/text supervision to further optimize v2q/v2t projection. Moreover, we devise a selection scheme to balance the two corrections. The whole architecture is optimized by minimizing query and text classification loss.

The main contribution of this work is the proposal of exploring the cross correction between query and text to boost weakly-supervised video representation learning. This also leads to a better view of why query and text could complement to each other to validate visual-textual connections, and how to integrate the correction across the two into video representation learning framework. Two large-scale web video datasets are proposed for the weakly-supervised learning and extensive experiments on the datasets demonstrate the effectiveness of our BCN framework.

\section{Related Work}

We briefly group the related works into three categories: supervised, unsupervised and weakly-supervised video representation learning with respect to the utilization of clean labels, no label or noisy labels for model training.

The early works \cite{Diba:CVPR17,Feichtenhofer:CVPR16,Sports1M,Yue-Hei:CVPR15,Simonyan:NIPS14,Wang:ECCV16} of supervised video representation learning are extended from image representation by applying 2D CNN on video frames. For instance, Karpathy \emph{et al.} \cite{Sports1M} leverage spatio-temporal convolutions for representation learning on the stacked frame-level features. To further capture the motion information, the well-known two-stream architecture \cite{Simonyan:NIPS14} and its variants \cite{Diba:CVPR17,Wang:ECCV16,TSN:PAMI} are devised by executing 2D CNN on optical flow.
Though the methods improve the representation learning by formulating motion pattern, performing 2D CNN on video frames still limits the capacity of modeling long-range temporal dynamics.
To alleviate this problem, LSTM-RNN \cite{Yue-Hei:CVPR15} captures the long-term dependencies in videos by utilizing a long-short term memory (LSTM) auto-encoder. To treat the video clip as a temporal evolution unit rather than a sequence of independent frames, 3D CNN structures \cite{Carreira:CVPR17,Hara:CVPR18,Qiu:ICCV17,Tran:ICCV15,Tran:CVPR18} are proposed to boost video feature learning on large-scale supervised video datasets \cite{Carreira:CVPR17,Kinetics:600}. 
More recently, the video transformer works \cite{Bertasius:ICML21,Fan:MVIT,ViViT} are extended from the vision transformers \cite{ViT,Swin-ViT} in image domain to learn video representation.
Note that the backbone of our BCN is 3D ConvNet but the representation is learnt in a weakly-supervised manner.

Unsupervised video representation learning is one kind of technique to employ unlabeled videos for representation learning.
The related works leverage various supervision from the video data itself to build pretext tasks for video representation learning, such as frame/clip order prediction \cite{Fernando:CVPR17,Misra:ECCV16,Wei:CVPR18,Xu:CVPR19}, motion estimation \cite{Agrawal:ICCV15,Pathak:CVPR17}, temporal cycle-consistency learning \cite{Dwibedi:CVPR19,Wang:CVPR19}, temporal coherence learning \cite{Mobahi:ICML09,Wang:ICCV15}, pixel-level displacement prediction \cite{Liu:ICCV17,Wang:CVPR19}, frame reconstruction \cite{Finn:NIPS16,Luo:CVPR17,Srivastava:ICML15} and contrastive learning \cite{Feichtenhofer:CVPR21,Li:ICLR21}.
To further improve the descriptive ability of video representation, weakly-supervised methods \cite{Ghadiyaram:CVPR19,Li:CPD20,Miech:CVPR20} focus on utilizing the weak supervision from web video data~\cite{Miech:ICCV19}.
For example, Ghadiyaram \emph{et al.} \cite{Ghadiyaram:CVPR19} explore the influences of different aspects in tags (e.g., label space) for weakly-supervised learning. Furthermore, to mine the knowledge from title of web videos, CPD \cite{Li:CPD20} learns the video representation by making visual-textual pair close to each other.
Nevertheless, most of the recent weakly-supervised methods are still facing the challenge of query polysemy or text isomorphism when directly exploiting query or text as supervision.
Particularly, some previous works \cite{Berg:CVPR06, Saenko:NIPS08, Schroff:ICCV07} handled query polysemy in web data collection, however our proposal alleviates it in the stage of representation learning.

In short, our approach belongs to weakly-supervised video representation learning techniques. Different from the aforementioned methods which solely rely on the supervision of query or text, our BCN in this paper contributes by studying not only mining supervisory signal from query and text simultaneously, but also how mutual calibration between query and text information could be leveraged to enhance weakly-supervised video representation learning.

\begin{figure*}[!tb]
	\centering
	\subcaptionbox{Feature learning on query supervision}{\label{fig1:2:a}\includegraphics[width=0.220\textwidth]{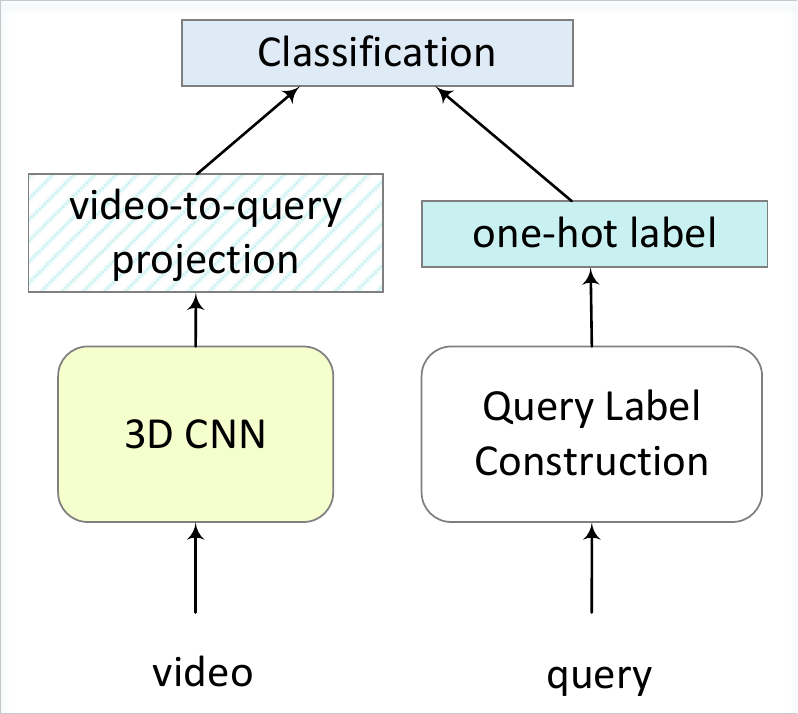}} \hspace{0.3in}
	\subcaptionbox{Feature learning on text supervision}{\label{fig1:2:b}\includegraphics[width=0.220\textwidth]{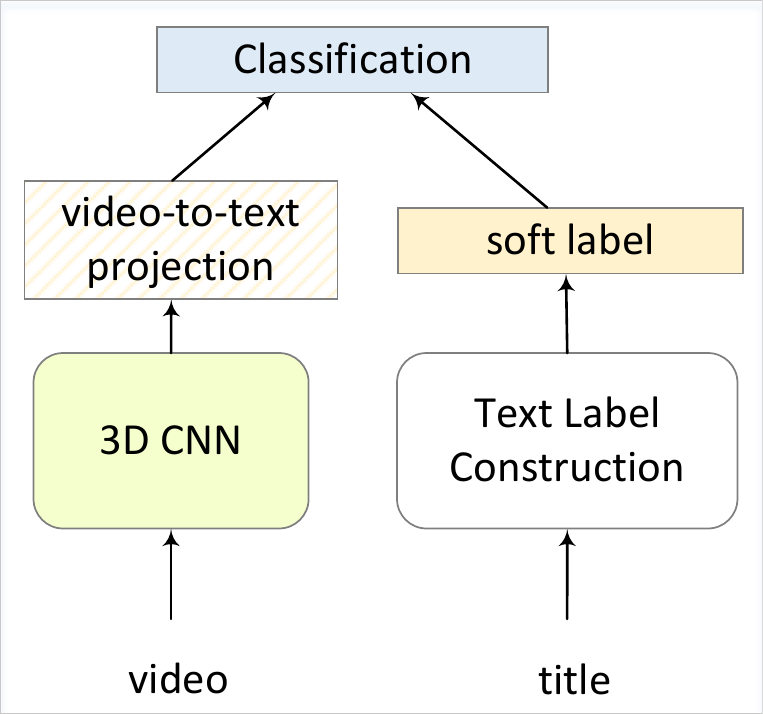}} \hspace{0.3in}
	\subcaptionbox{Feature learning on query and text supervision through bi-directional calibration}{\label{fig1:2:c}\includegraphics[width=0.420\textwidth]{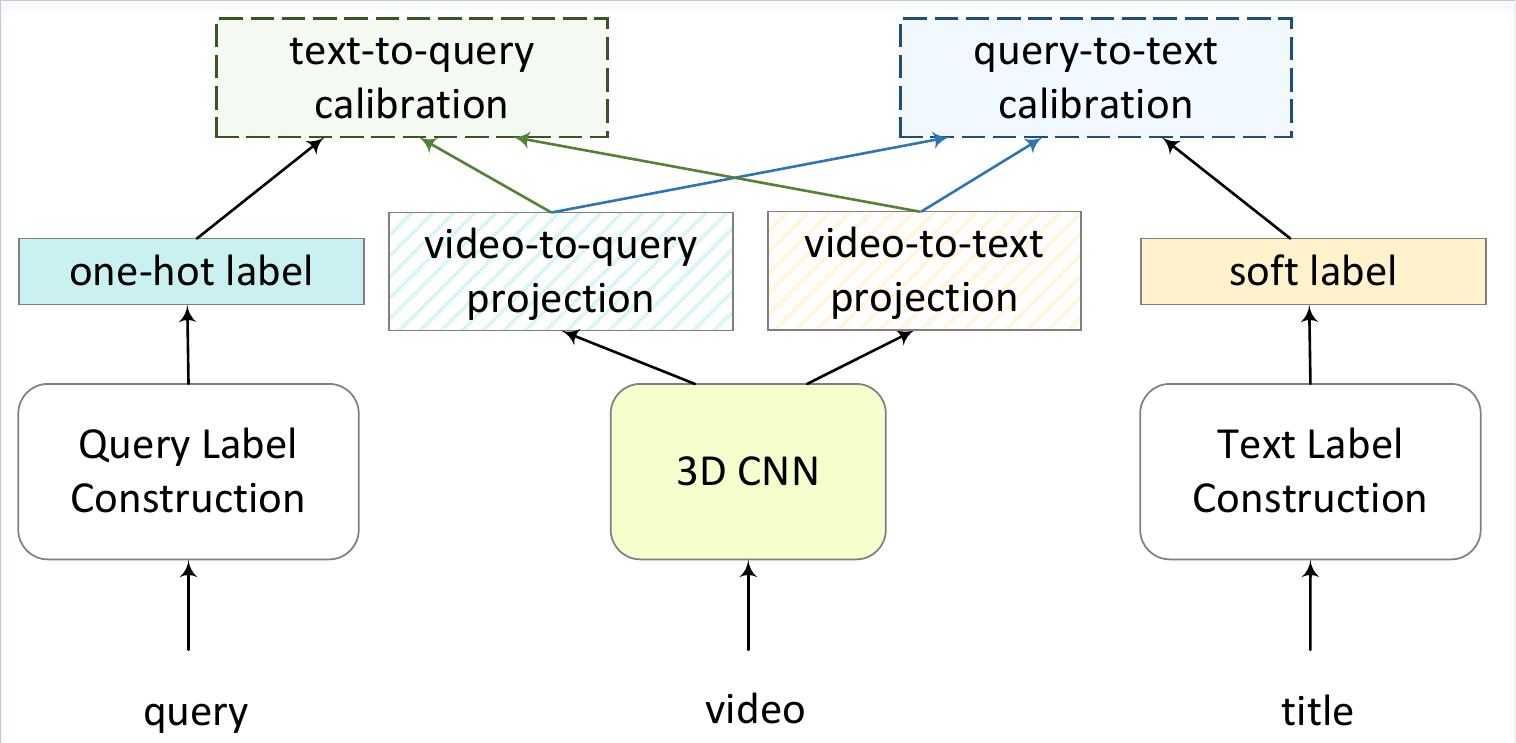}}
	\vspace{-0.01in}
	\caption{\small The conceptual illustration of weakly-supervised video representation learning on (a) query supervision; (b) text supervision; (c) both of query and text supervision with bi-calibration in BCN.}
	\label{fig:1:2}
	\vspace{-0.1in}
\end{figure*}

\section{Bi-Calibration Networks}
In this section, we introduce the Bi-Calibration Networks (BCN) that performs mutual calibration between query and text to facilitate weakly-supervised video representation learning. Specifically, BCN formulates the problem as the classification over query and text vocabulary. To better understand the spirit of our BCN design, we first introduce the problem formulation of the unique pretext task in BCN for weakly-supervised video representation learning. After that, the detailed BCN architecture is further elaborated.   

\subsection{Problem Formulation}
Suppose we have a web video collection, where each video is paired with the searched query and the video title description.
The goal of the weakly-supervised video representation learning is to pre-train a visual encoder on the video data supervised by the query or text supervision.
The pre-trained visual encoder can be further utilized to support several downstream tasks.
In this paper, we employ the 3D CNN as the visual encoder for video pre-training.

For the video pre-training base on query, the query words are usually converted to ``one-hot'' labels.  
One natural way \cite{Ghadiyaram:CVPR19} to optimize the 3D CNN is to classify the video-to-query (v2q) projection of the 3D backbone based on these semantic labels, as shown in Figure \ref{fig:1:2}(a).
On the other hand, there are many directions to employ text to supervise video representation learning, such as the contrastive learning \cite{Li:CPD20}, triplet ranking \cite{Stroud:Trp20} or linear feature regression \cite{Qiu:CVPR21}. 
Instead, we still formulate the weakly-supervised video feature learning on text as the classification pretext task.
As illustrated by Figure \ref{fig:1:2}(b), we convert the video titles into the soft labels and further employ them as the text supervision to optimize the 3D CNN through the classification on video-to-text (v2t) projection.  
To improve the quality of the primary query/text supervision and facilitate video representation learning, we design the bi-directional calibration learning paradigm to correct query and text supervision across each. The text-to-query (t2q) or query-to-text (q2t) calibration takes both of the v2q and v2t projections as the correction signals to refine the primary ``one-hot'' query label or "soft" text label, as illustrated by Figure \ref{fig:1:2}(c).

\begin{figure*}[!tb]
	\centering\includegraphics[width=0.99\textwidth]{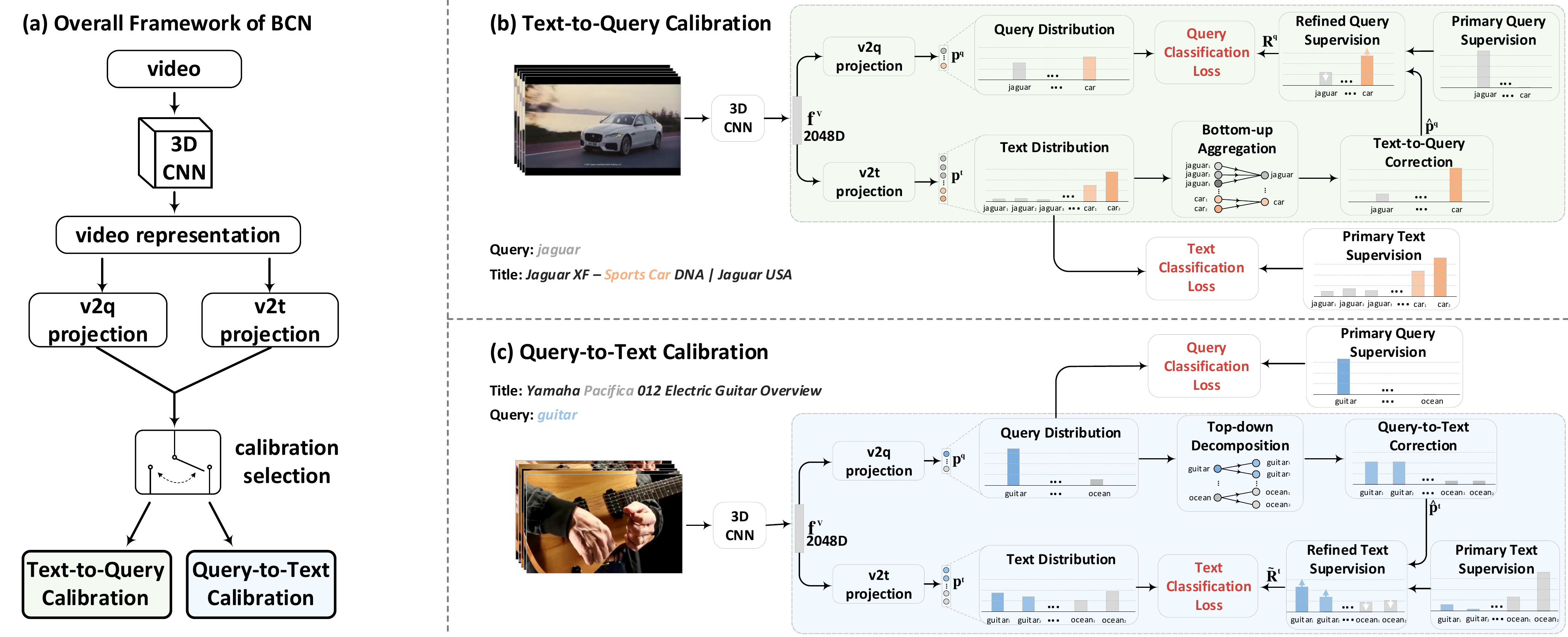}
	\caption{\small
		An overview of our Bi-Calibration Networks (BCN). In general, BCN (a) first extracts the video representation of input video via 3D CNN, and then feed it into two branches, i.e., video-to-query (v2q) and video-to-text (v2t) projections. The output query distribution over query vocabulary and the text distribution over all text prototypes are utilized to trigger the text-to-query (b) and query-to-text (c) calibration modules, which are controlled by a selection scheme.}
	\label{fig2:1}
\end{figure*}

Figure \ref{fig2:1} further details an overview of the BCN framework and two coupled calibration modules, i.e., text-to-query (t2q) and query-to-text (q2t) calibration. Specifically, for each input video, BCN first utilizes a 3D CNN to extract video representation and feed it into two branches, i.e., video-to-query (v2q) and video-to-text (v2t) projections. Two kinds of probability distributions (i.e., the query distribution over query vocabulary and the text distribution over all text prototypes) are achieved to trigger each calibration. Note that all the titles of the videos searched by an identical query are initially grouped into multiple clusters, and each text prototype corresponds to the centroid of each cluster. In this way, we naturally obtain two kinds of primary supervision for each video, i.e, the primary query supervision (the ``one-hot'' vector in query vocabulary) and the primary text supervision (the cosine similarity between video title and all text prototypes), that are used to optimize v2q and v2t projections, respectively. After that, BCN starts t2q/q2t calibration by aggregating/decomposing the text/query distribution into t2q/q2t correction in a bottom-up/top-down fashion, respectively. The learnt t2q/q2t correction is further integrated with the primary query/text supervision, yielding the refined query/text supervision to strengthen the regulation of each branch. During training, a selection scheme is utilized to balance the two calibrations.

\begin{figure}[!tb]
	\centering\includegraphics[width=0.49\textwidth]{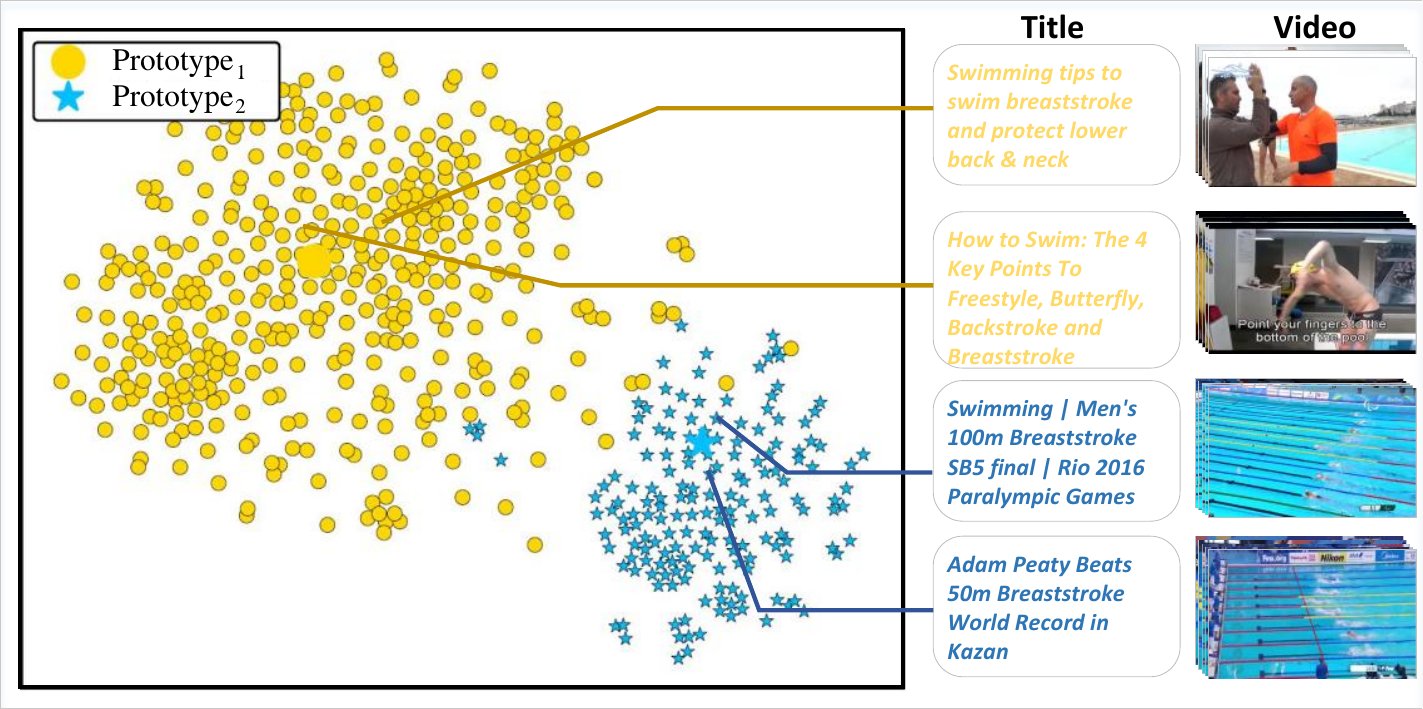}
	\vspace{-0.02in}
	\caption{\small Visualization of clustering on title features by t-SNE \cite{Maaten:JMLR8}. (query: swimming breast stroke)}
	\label{fig2:2}
	\vspace{-0.15in}
\end{figure}

\subsection{Video-to-Query/Text Projection Branch}
The ultimate target of BCN is to train the 3D CNN backbone for video representation learning through two pretext tasks of query and text classification. Therefore, based on the extracted video feature, we involve two parallel video-to-query (v2q) and video-to-text (v2t) projection branches to perform the two pretext tasks, which can be optimized with the corresponding query and text supervision.

\textbf{Primary Query and Text Supervision.}
Given all the training web videos paired with the searched queries and surrounding text (i.e., titles), one natural way to represent each query or text as query/text supervision is to directly construct query/text vocabulary based on query/title words. Therefore, for each query, we adopt this natural way to take its ``one-hot'' vector $\mathbf{y^q} \in {\mathbb{R}}^K$ in query vocabulary (vocabulary size: $K$) as the primary query supervision. However, the robustness of this recipe is brittle when applying it to represent surrounding texts, since the inherently semantic distribution among different titles is unexploited. Inspired by the classical bag-of-words paradigm for feature representation \cite{Lazebnik:CVPR06}, we leverage such paradigm to construct text vocabulary by delving into the diverse semantic structure of titles. Formally, we first extract the text feature $\mathbf{f^t}$ of each video title by the off-the-shelf language model BERT \cite{Jacob:ACL18}. Next, for the $k$-th query, we perform clustering on all the titles of the videos searched by this query through k-means algorithm, leading to $m_k$ clusters. The cluster number $m_k$ is automatically set by a statistic-based clustering estimation method (Gap Statistic \cite{Tibshirani:Royal01}). The centroid $\mathbf{B}_i$ of the $i$-th cluster is then defined as one text prototype, which is computed as the average of all text features of video titles in that cluster. Accordingly, we construct the text vocabulary based on all the $M=\sum_{k=0}^{K-1}{m_k}$ text prototypes derived from $K$ queries, which naturally characterize the underlying semantic structure of titles. Based on this text vocabulary, we interpret the primary text supervision $\mathbf{y^t} = [y^t_0, y^t_1, ..., y^t_{M-1}] \in {\mathbb{R}}^M$ of each surrounding text $\mathbf{f^t}$ as the cosine similarities between $\mathbf{f^t}$ and all the $M$ text prototypes $\{\mathbf{B}_i|{i=0,1,...,M-1}\}$:
\begin{equation}\label{Eq2:1}\small
	\mathbf{y^t} = softmax(COS(\mathbf{f^t},\mathbf{B}_i)|i=0,1,...,M-1), \\
\end{equation}
where $softmax(\cdot)$ is the softmax with temperature parameter $1/M$ and $COS(\cdot)$ represents cosine similarity. 

One case of the clustering for video titles searched by the query of ``swimming breast stroke'' is visualized in Figure \ref{fig2:2}. There are two meaningful clusters as shown in the figure. The first cluster has an emphasis on the daily breaststroke (e.g., tutorial) while most of the titles in the second cluster tend to describe swimeet in the championship. The differences in between are reflected in visual content. Thus, the text supervision based on the similarity between title feature and each prototype well captures the semantic relation.

\textbf{Video-to-Query Projection.}
The v2q projection branch is especially designed to perform the pretext task of query classification. Concretely, depending on the input video feature $\mathbf{f^v}$ extracted by 3D CNN, the v2q projection branch infers its query distribution $\mathbf{p^q}=[p^q_0, p^q_1,...,p^q_{K-1}]$ over query vocabulary through a fully-connected layer. Note that $p^q_k$ is the estimated probability of assigning this video to the $k$-th query word. The v2q projection branch can be directly optimized with the primary query supervision $\mathbf{y^q}$, and we measure the query classification loss as softmax loss:
\begin{equation}\label{Eq2:3}\small
	L^q = -\sum_{k=0}^{K-1}{I_{k=y^q}\log(p^q_k)},  \\
\end{equation}
where the indicator function $I_{k=y^q}=1$ if the label value of $k$-th query word in $\mathbf{y^q}$ is 1, otherwise $I_{k=y^q}=0$.

\textbf{Video-to-Text Projection.}
In analogy to the v2q projection branch, we design another v2t projection branch to conduct the pretext task of text classification. In particular, the v2t projection branch takes the video feature $\mathbf{f^v}$ as the input, and learns to estimate the text distribution $\mathbf{p^t}=[p^t_0, p^t_1,...,p^t_{M-1}]$ over all text prototypes in text vocabulary. $p^t_i$ denotes the probability of the video belonging to the $i$-th cluster. Accordingly, we optimize the v2t projection branch with the primary text supervision $\mathbf{y^t} = [y^t_0, y^t_1, ..., y^t_{M-1}]$, and the text classification loss for this proxy task is calculated as the softmax cross-entropy loss:
\begin{equation}\label{Eq2:2}\small
	L^t = -\sum_{i=0}^{M-1}{y^t_i\log(p^t_i)}.  \\
\end{equation}

\subsection{Text-to-Query Calibration}

The most typical way to optimize video-to-query projection branch is to use the primary query supervision for query classification as in Eq.(\ref{Eq2:3}). However, such way oversimplifies the proxy task by assuming that all videos searched by an identical query belong to one class, while ignoring the phenomenon of query polysemy (i.e., the coexistence of many possible meanings for a query). That will inevitably mislead video representation learning. To alleviate this problem, we devise a text-to-query (t2q) calibration module that further regulates the v2q projection with additional calibration from video surrounding texts (i.e., titles).

\textbf{Text-to-Query Correction.}
The t2q calibration module first transforms the text distribution $\mathbf{p^t}$ derived from the v2t projection branch into the t2q correction $\mathbf{\hat{p}^q} \in {\mathbb{R}}^K$ by aggregating all the probabilities of text prototypes belonging to the same query in a bottom-up way:
\begin{equation}\label{Eq2:4}\small
	\begin{split}
		&\hat{p}^q_k = \sum_{i \in I_k}{p^t_i}, ~~~s.t.~~~  k \in \{0,1,...,K-1\},
	\end{split}
\end{equation}
where $\hat{p}^q_k$ denotes the $k$-th correction value in t2q correction $\mathbf{\hat{p}^q}$ and represents the aggregated probability with regard to the $k$-th query word in query vocabulary. $I_k$ is the index set of the text prototypes for the $k$-th query.

\textbf{Refined Query Supervision.}
Next, the t2q correction $\mathbf{\hat{p}^q}$ serves as the additional calibration from the text distribution over all text prototypes, aiming to refine the primary query supervision $\mathbf{y^q}$ with more semantic meanings derived from v2t projection branch. Specifically, we first take the primary query supervision $\mathbf{y^q}$ weighted by the estimated query distribution $\mathbf{p^q}$ as the query confidence $\mathbf{y^{q*}}= \mathbf{y^q} \circ \mathbf{p^q}$, which reflects the confidence score of each query word. Note that the operation $\circ$ denotes the element-wise multiplication. The refined query supervision $\mathbf{R^q}$ is then estimated by integrating the t2q correction $\mathbf{\hat{p}^q}$ with the query confidence score $\mathbf{y^{q*}}$ as:
\begin{equation}\label{Eq2:5}\small
	\mathbf{R^q} =  (\mathbf{{y}^{q*}} + \mathbf{\hat{p}^q}) / \|\mathbf{{y}^{q*}} + \mathbf{\hat{p}^q}\|_1,\\
\end{equation}
where $\|\cdot\|_1$ represents $\mathcal{L}_1$-norm. The underlying assumption behind Eq.(\ref{Eq2:5}) is that if the confidence score of the ground-truth query word is higher than the correction value of that query, the refined query supervision prefers to be closer to the primary query supervision. Otherwise, the refined query supervision tends to be heavily influenced with the t2q correction.
Finally, the t2q calibration module leverages the refined query supervision $\mathbf{R^q}$ to further optimize the v2q projection branch, and the query classification loss in Eq.(\ref{Eq2:3}) is thus reformulated as:
\begin{equation}\label{Eq2:6}\small
	\hat{L}^q = -\sum_{k=0}^{K-1}{R^q_{k}\log(p^q_k)},  \\
	\vspace{-0.05in}
\end{equation}
where $R^q_{k}$ is the updated ground-truth of $k$-th query word in the refined query supervision $\mathbf{R^q}$.

\subsection{Query-to-Text Calibration}

Recall that the aforementioned optimization of the v2t projection branch (see Eq.(\ref{Eq2:2})) solely hinges on the primary text supervision for text classification proxy task. Nevertheless, in the case of text isomorphism (i.e., titles share the same syntactic structure but refer to different semantics), the semantic discriminativeness of video representations learnt in this way may be easily overwhelmed. To address the issue, a query-to-text (q2t) calibration module is designed to guide the optimization of v2t projection with the additional high-level semantic supervision from query.

\textbf{Query-to-Text Correction.} In the q2t calibration module, we first calculate the q2t correction $\mathbf{\hat{p}^t} \in {\mathbb{R}}^M$ by evenly decomposing each element (e.g., the probability $p^q_k$ of $k$-th query word) of query distribution to the correction values over all the text prototypes belonging to that query:
\begin{equation}\label{Eq2:7}\small
	\begin{split}
		\hat{p}^t_i = \frac{1}{m_k} {p^q_k}, ~~~s.t.~~~ i \in I_k,~~ k \in \{0,1,..., K-1\},
	\end{split}
\end{equation}
where $\hat{p}^t_i$ denotes the correction value of the $i$-th text prototype in q2t correction $\mathbf{\hat{p}^t}$ and $m_k$ is the number of text prototypes belonging to the $k$-th query.

\textbf{Refined Text Supervision.} After that, we improve the primary text supervision with the q2t correction, leading to the refined text supervision to further regulate the v2t projection branch. In particular, the text confidence $\mathbf{y^{t*} }= \mathbf{y^t} \circ \mathbf{p^t}$ is first calculated as the primary text supervision $\mathbf{y^t}$ weighted by text distribution $\mathbf{p^t}$. Each element in $\mathbf{y^{t*}}$ denotes the confidence score of each text prototype. Similar to the formulation of refined query supervision, we measure the refined text supervision $\mathbf{R^t}$ by fusing the q2t correction $\mathbf{\hat{p}^t}$ and the text confidence $\mathbf{y^{t*}}$:
\begin{equation}\label{Eq2:8}\small
	\mathbf{R^t} =  (\mathbf{{y}^{t*}} + \mathbf{\hat{p}^t}) / \|\mathbf{{y}^{t*}} + \mathbf{\hat{p}^t}\|_1.\\
\end{equation}
Moreover, in order to make the refined text supervision evolve smoothly, we utilize a moving average update strategy with momentum to update the refined text supervision:
\begin{equation}\label{Eq2:9}\small
	\mathbf{\widetilde{R}^{t}} \leftarrow \alpha\mathbf{\widetilde{R}^{t}} + (1-\alpha)\mathbf{{R}^{t}},
\end{equation}
where $\alpha$ is a momentum coefficient, and $\mathbf{R^{t}}$ or $\mathbf{\widetilde{R}^{t}}$ denotes the new or running value of the refined text supervision, respectively. Based on the updated refined text supervision $\mathbf{\widetilde{R}^t}$, the q2t calibration module further optimizes the v2t projection branch through the text classification loss:
\begin{equation}\label{Eq2:10}\small
	\hat{L}^t = -\sum_{i=0}^{M-1}{\widetilde{R}^t_{i}\log(p^t_i)},  \\
\end{equation}
where $\widetilde{R}^t_{i}$ is the new ground-truth of the $i$-th text prototype in the updated refined text supervision $\mathbf{\widetilde{R}^t}$.

\subsection{Network Optimization} \label{sec:3.4}
During training, we adopt a two-stage strategy to optimize the whole architecture of our BCN. In the first stage, we optimize the BCN framework with the typical query and text classification losses ($L^q$ in Eq.(\ref{Eq2:3}) and $L^t$ in Eq.(\ref{Eq2:2})) simultaneously, irrespective of any calibration modules. In the second stage, the BCN framework is further fine-tuned with two coupled calibration modules. Here, we design a selection scheme to balance the two calibration modules according to the difference between the query/text confidence ($\mathbf{{y}^{q*}}$/$\mathbf{{y}^{t*}}$) and t2q/q2t correction ($\mathbf{\hat{p}^q}$/$\mathbf{\hat{p}^t}$) in each module.
\begin{table*}[!tb]
	\centering
	\caption{\small The statistics of YOVO-3M and YOVO-10M datasets.}
	\vspace{0.0in}
	\scalebox{0.98}[0.98]{
		\begin{tabular}{{c|c c c c c c}}
			\hline
			Dataset  & ~~~~Source~~~  & Supervision &  ~~~\# of Video~~~ & ~~~\# of Clip~~~ & ~~\# of Query~~ & Duration (hrs)  \\ \hline \hline
			YOVO-3M  & web data  & query and title & 986,031   & 2,958,092  & 2,015  & 8216.9 \\
			YOVO-10M & web data  & query and title & 8,051,431 & 10,023,532 & 18,305 & 12142.1 \\ \hhline{*{7}{-}}
		\end{tabular}
	}
	\vspace{-0.0in}
	\label{table:datasets}
\end{table*}
\begin{algorithm}[htb]\small
	\SetKwInOut{Input}{Input}
	\SetKwInOut{Output}{Output}
	\caption{\small Calibration Selection Scheme}
	\label{CSS}
	{\fontsize{9}{9}\selectfont
		\Input{
			Model $\mathbf{M}$ of the first training stage, web videos with query and title; two thresholds $\varepsilon^q$ and $\varepsilon^t$;}
		\Output{
			Output model $\mathbf{\hat{M}}$ of BCN;}
		Initialize BCN model with $\mathbf{M}$, the iterative count $n=0$;\\
		\While{$n \leq N-1$}{
			Network forward to obtain probabilities $\mathbf{p^q}$ and $\mathbf{p^t}$; \\
			Compute $\mathbf{\hat{p}^q}$ and $\mathbf{\hat{p}^t}$ according to Eq.(\ref{Eq2:4}) and Eq.(\ref{Eq2:7});\\
			Compute $\mathbf{y^{q*}}$, $\mathbf{R^q}$ and $\mathbf{y^{t*}}$, $\mathbf{\widetilde{R}^t}$ by Eq.(\ref{Eq2:5}) and Eq.(\ref{Eq2:8}), Eq.(\ref{Eq2:9});\\
			\uIf{$ \|\mathbf{y^{q*}}-\mathbf{\hat{p}^q}\|_2 > \varepsilon^q $ \textbf{and} $\|\mathbf{y^{t*}}-\mathbf{\hat{p}^t}\|_2 < \varepsilon^t$}{
				Optimize $\mathbf{\hat{M}}$ by $\hat{L}^q$ and ${L}^t$; (t2q calibration)}
			\uElseIf{$\|\mathbf{y^{t*}}-\mathbf{\hat{p}^t}\|_2 > \varepsilon^t$ \textbf{and} $ \|\mathbf{y^{q*}}-\mathbf{\hat{p}^q}\|_2 < \varepsilon^q$}{
				Optimize $\mathbf{\hat{M}}$ by $\hat{L}^t$ and ${L}^q$; (q2t calibration)}
			\lElse{Optimize $\mathbf{\hat{M}}$ by ${L}^q$ and ${L}^t$}
			$n = n + 1$;}
		\Return $\mathbf{\hat{M}}$}
\end{algorithm}
Most specifically, if $\small{\|\mathbf{y^{q*}}-\mathbf{\hat{p}^q}\|_2 > \varepsilon^q}$ and $\small{\|\mathbf{y^{t*}}-\mathbf{\hat{p}^t}\|_2 < \varepsilon^t}$ ($\varepsilon^q$ and $\varepsilon^t$ are two thresholds), this case implies that the primary text supervision ($\mathbf{y^{t}}$) is more reliable than the primary query supervision ($\mathbf{y^{q}}$). Therefore we select the t2q calibration module, and the BCN framework is optimized with the modified query classification loss $\hat{L}^q$ in Eq.(\ref{Eq2:6}) plus the typical text classification loss $L^t$ in Eq.(\ref{Eq2:2}).
In contrast, if $\small{\|\mathbf{y^{q*}}-\mathbf{\hat{p}^q}\|_2 < \varepsilon^q}$ and $\small{\|\mathbf{y^{t*}}-\mathbf{\hat{p}^t}\|_2 > \varepsilon^t}$, the primary query supervision is supposed to be more reliable than the primary text supervision, and the q2t calibration module is selected for optimization. In that case, the objective of BCN framework consists of the typical query classification loss $L^q$ in Eq.(\ref{Eq2:3}) and the modified text classification loss $\hat{L}^t$ in Eq.(\ref{Eq2:10}). Otherwise, we optimize BCN with the typical query and text classification losses as in the first stage.
The weight for each loss is set as $1.0$ empirically.

Algorithm \ref{CSS} details the processing of the selection scheme of the two calibrations in our BCN.

\section{Experiments}
The experiments of weakly-supervised video representation learning are conducted on the newly-created YouTube video datasets, namely YOVO-3M and YOVO-10M, particularly paired with query and title. We then empirically verify the merit of BCN on three scene-related action recognition datasets: Kinetics-400 \cite{Carreira:CVPR17}, UCF101 \cite{Soomro:UCF101} and HMDB51 \cite{Kuehne:ICCV11}, and two interaction-related action recognition datasets: Something-Something V1 and V2 \cite{Goyal:SS}.

\subsection{Datasets}
\begin{figure}[!tb]
	\centering\includegraphics[width=0.49\textwidth]{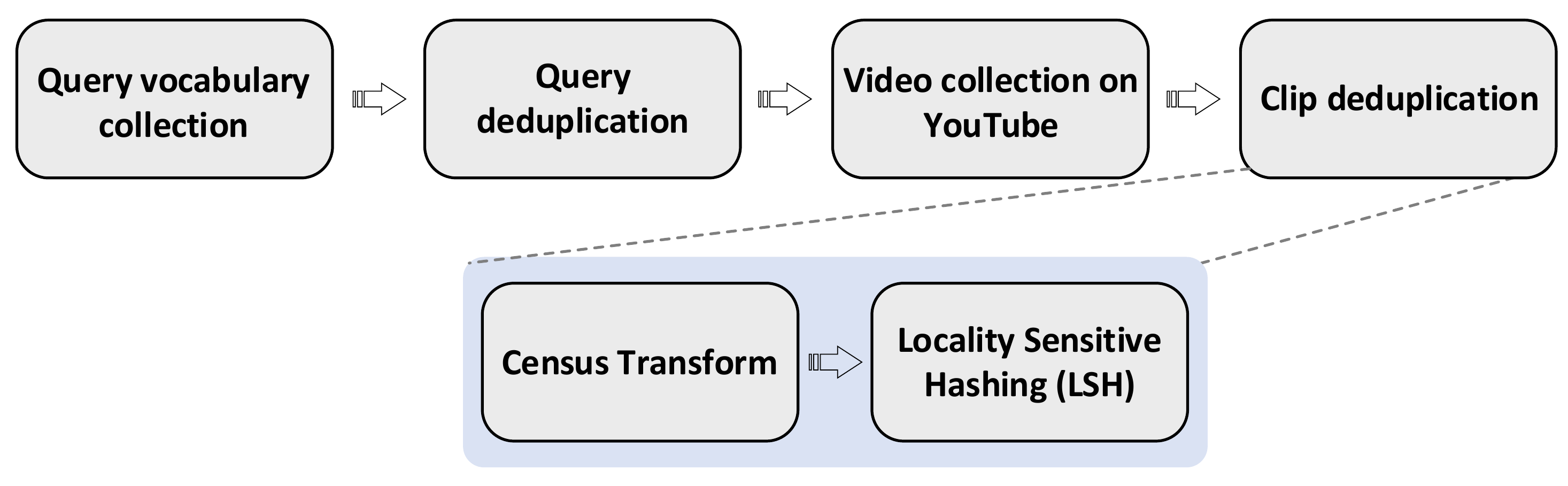}
	\vspace{-0.1in}
	\caption{\small The construction pipeline of YOVO-3M and YOVO-10M.}
	\label{fig:dataset}
	\vspace{-0.1in}
\end{figure}

\textbf{YOVO-3M/10M Datasets.} 
We collect the YOVO-3M and YOVO-10M datasets characterized by the unique properties including large-scale web video data with query and title information, as well as the comprehensive and diverse video content for weakly-supervised video representation learning. Figure \ref{fig:dataset} depicts the construction pipeline of YOVO-3M and YOVO-10M datasets, which consists of four main steps, i.e., query vocabulary collection, query deduplication, video collection on YouTube and clip deduplication.
To crawl the web videos with comprehensive visual content, we first collect all the labels from Kinetics-400 \cite{Carreira:CVPR17}, Kinetics-700 \cite{Kinetics700}, ImageNet \cite{ImageNetDatabase} and Moments \cite{Moments} datasets as search queries. After query deduplication, the number of remaining queries is 2,015. We issue each query to YouTube and about 489 videos with the titles are downloaded successfully on average. We then uniformly sample three 10-second clips from each video to build YOVO-3M dataset. To further enlarge the volume of web videos, a 3K verb vocabulary and a 13K noun vocabulary extracted from Oxford English Dictionary are utilized as additional queries to search for another set of video clips with the titles. We combine these clips with YOVO-3M dataset to construct YOVO-10M dataset. Finally, we employ the standard clip deduplication approach \cite{Ghadiyaram:CVPR19} to remove video clips occurring anywhere in the downstream datasets from both YOVO-3M and YOVO-10M datasets. Specifically, we extract the global feature of each frame in a video clip through Census Transform \cite{Zasbih:ECCV94} and then execute Locality Sensitive Hashing (LSH) \cite{Gionis:VLDB99} on the feature to obtain frame-level hash codes. We then compute the hamming distance between the frame-level hash codes of each pair of frames, in which one is from a video clip from YOVO-3M/10M datasets and the other is from a video in downstream datasets, e.g., Kinetics-400. We average all the distances of frame pairs across two videos as the clip-level distance in between. If the distance is lower than $2$, we will remove the corresponding video clip from YOVO-3M/10M.

\begin{figure*}[!tb]
	\centering\includegraphics[width=1.0\textwidth]{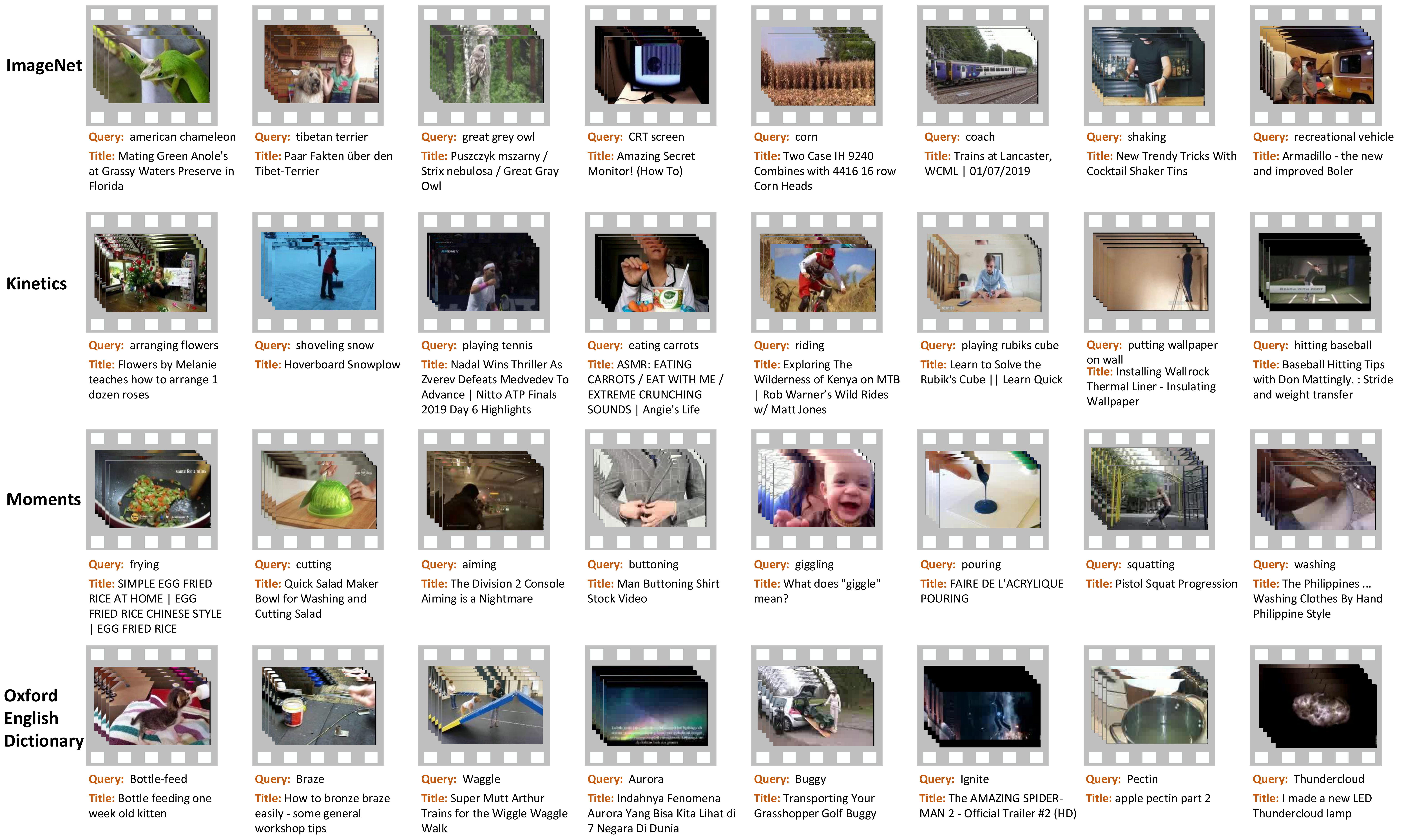}
	\caption{\small Examples of 32 video clips with searched queries and titles from YOVO-3M and YOVO-10M datasets. The video clips in each row are searched by the queries in the category list of different datasets, i.e., ImageNet, Kinetics and Moments, or in the vocabulary list of the Oxford English Dictionary.}
	\label{fig:yovo-samples}
	\vspace{-0.1in}
\end{figure*}

Table \ref{table:datasets} summarizes the statistics of YOVO-3M and YOVO-10M datasets. In detail, YOVO-3M contains $2,015$ queries and $2,958,092$ video clips in total, and YOVO-10M consists of $18,305$ queries and $10,023,532$ video clips. 
The two scales of the proposed datasets would be applicable to different researchers w.r.t computational resources in the video representation learning community. 
Figure \ref{fig:yovo-samples} further illustrates $32$ video clips with the searched queries and titles from YOVO-3M and YOVO-10M datasets.
The showing video cases demonstrate the diverse video content in different facets, e.g., objects, sports and daily activities, for weakly-supervised video representation learning.

\textbf{Downstream Datasets.} We evaluate our BCN on five downstream datasets. The Kinetics-400 \cite{Carreira:CVPR17} dataset consists of 300K 10-second clips from 400 action categories. There are 240K, 20K, 40K clips in training, validation and testing sets, respectively. The UCF101 \cite{Soomro:UCF101} contains 13K videos from 101 action classes, and the HMDB51 \cite{Kuehne:ICCV11} has 7K videos from 51 action categories.
In UCF101 and HMDB51 datasets, there are three training/validation splits provided by the dataset organizers. Each split in UCF101 includes about 9.5K training and 3.7K validation videos, and a HMDB51 split contains 3.5K training and 1.5K validation videos. The videos in the above three datasets are mainly for scene-related action recognition. In Something-Something V1 and V2 \cite{Goyal:SS} datasets, there are about 108K and 221K videos from 174 action categories, respectively. The training/validation/testing set consists of 86K/11.5K/11K and 169K/25K/27K videos, which are mostly for interaction-related recognition.

\subsection{Experimental Settings}

\textbf{Implementations.} We employ BERT \cite{Jacob:ACL18} text encoder to extract the 1,024-D features for each token in the title and average all the token features as text representation. The number $M$ of total clusters is $6,819$/$65,766$ on YOVO-3M/10M. In weakly-supervised training, we utilize the architecture of LGD-3D \cite{Qiu:CVPR19} based on ResNet-50 \cite{Kaiming:CVPR16} as the network backbone but all the parameters are trained from scratch. The dimension of the input video clips is set as $32\times 112\times 112$, which is randomly cropped from the original web video. Each clip is randomly flipped along horizontal direction for data augmentation.
We choose the two threshold $\varepsilon^q$ and $\varepsilon^t$ as $0.5$ and $0.7$ by cross validation.
The momentum coefficient $\alpha$ is fixed to $0.9$. We implement BCN on Caffe \cite{Yang:caffe} platform. In all the pre-training stages, the networks are trained by utilizing stochastic gradient descent (SGD) with $0.9$ momentum. The initial learning rate is set to $0.08$ and $0.008$ in the first and second training stage, and decreased by $10\%$ after every $200$K iterations. The mini-batch size is set as $256$ and the weight decay is $0.0001$.

\textbf{Evaluation Metrics.} We adopt two evaluation protocols in the downstream datasets, i.e., linear model and network fine-tuning. In the former protocol, we uniformly sample 10 or 3 video clips from each video in Kinetics-400/UCF101/HMDB51 or Something-Something V1/V2 datasets, and take the $2,048$-way outputs from pool5 layer of the network backbone as the features of each clip. We average all the features of clips in one video as video representation, and a linear SVM is learnt on the training set and evaluated on validation set. In detail, the cost parameter $c$ in SVM is set as $8.0$. Both top-1 and top-5 accuracy are reported as evaluation metrics. In the latter one, we initialize the network backbone with the weakly-supervised training output model of BCN, and fine-tune/evaluate the network on the training/validation set of each dataset.

\subsection{Evaluation on Primary Text Supervision}
We first examine the effectiveness of primary text supervision for video representation learning, regardless of mutual calibration design. We compare the following four methods: (1) The regression method (RG) optimizes video representation through minimizing Smooth L1 loss \cite{Girshick:ICCV15} between video representation and text feature. (2) Triplet Ranking algorithm (TR) learns video representation to make positive video-text pair more similar than negative pair. (3) A variant of our primary text supervision (TS-) also performs clustering on video titles and builds text vocabulary on the centroids of clusters. Each title is naturally assigned to one cluster and represented as a binary index vector in text vocabulary. We exploit single-label classification to regulate video representation with text supervision in TS-. (4) TS is our proposed primary text supervision in BCN, which computes cosine similarity between title and all clusters. TS can be regarded as the soft mode of TS- and accordingly tunes video feature with multi-label classification loss.

\begin{table}[!tb]
	\setlength{\belowcaptionskip}{-1pt}
	\centering
	\caption{\small Top-1 and Top-5 accuracy on Kinetics-400, UCF101 and HMDB51 under linear protocol. (Training on YOVO-3M).}
	\scalebox{0.85}[0.85]{
		\begin{tabular}{{|c|c@{~}c|c@{~}c|c@{~}c|}}
			\hline
			\multicolumn{1}{|c|}{\multirow{2}{*}{\text{Approach}}} & \multicolumn{2}{c|}{\text{Kinetics-400}} & \multicolumn{2}{c|}{\text{UCF101}} & \multicolumn{2}{c|}{\text{HMDB51}}\\ \hhline{*{1}{~}*{6}{-}}
			\multicolumn{1}{|c|}{} & ~Top-1~~ & ~Top-5~~ & ~Top-1~~ & ~Top-5~~ & ~Top-1~~ & ~Top-5~~ \\ \hhline{*{7}{-}}
			RG     & 64.5 & 84.9 & 85.1 & 94.3  & 58.3  & 81.2 \\
			TR     & 67.2 & 87.2 & 87.2 & 95.7  & 61.8  & 83.9  \\
			TS-    & 71.5 & 89.1 & 91.0 & 98.9  & 62.2  & 85.6 \\ \hhline{*{7}{-}}
			TS     & 72.5 & 89.6 & 91.8 & 99.0  & 64.5  & 89.7  \\ \hhline{*{7}{-}}
		\end{tabular}
	}
	\label{table4:1}
\end{table}

\begin{table}[!tb]
	\setlength{\belowcaptionskip}{-1pt}
	\centering
	\caption{\small Performance contribution of each design in BCN. The model is learnt on YOVO-3M and evaluated by linear protocol. The ``Query'' and ``Text'' denote the corresponding supervision.}
	\scalebox{0.88}[0.88]{
		\begin{tabular}{{l|c c c c|c|c|c}}
			\hline
			\multicolumn{1}{c|}{\multirow{1}{*}{\text{Method}}} & Query & Text & T$\rightarrow$Q &  Q$\rightarrow$T  & K400 & U101 & HD51  \\ \hhline{*{8}{-}}
			QS                & \checkmark  &            &             &             & 71.1 & 90.7 & 62.1  \\
			TS                &             & \checkmark &             &             & 72.5 & 91.8 & 64.5 \\
			QS+TS             & \checkmark  & \checkmark &             &             & 73.1 & 92.3 & 65.9 \\
			BCN$_Q$           & \checkmark  & \checkmark & \checkmark  &             & 73.8 & 93.3 & 66.9 \\
			BCN$_T$           & \checkmark  & \checkmark &             & \checkmark  & 73.6 & 92.8 & 66.3 \\ \hline \hline
			\text{BCN}        & \checkmark  & \checkmark & \checkmark  & \checkmark  & \textbf{74.2} & \textbf{93.5} & \textbf{67.6} \\ \hhline{*{8}{-}}
		\end{tabular}
	}
	\label{table4:2}
	\vspace{-0.0in}
\end{table}

Table \ref{table4:1} summarizes performance comparisons of video representation learnt with different ways of primary text supervision under linear protocol on three downstream datasets. Overall, the results across different datasets consistently indicate that TS leads to a performance boost against other methods. In particular, the top-1 accuracy of our TS achieves 72.5\%, 91.8\% and 64.5\% on Kinetics, UCF101 and HMDB51, respectively, making the absolute improvement over RG/TR by 8.0\%/5.3\%, 6.7\%/4.6\% and 6.2\%/2.7\%. Such results demonstrate the advantage of exploring the structure among all the titles of videos via clustering. Though both TS- and TS utilize text clustering to improve text supervision, they are different in the way that TS- represents each title as an index vector (1 for its own cluster, otherwise 0), and TS is by computing cosine similarity between the title and all clusters. As indicated by the results, delving into the correlation between each title and all title clusters in TS leads to better performances.

\textbf{Discussion with Contrastive Learning.} Inspired from the self-supervised learning \cite{He:CVPR20}, the video representation learning from text is recently formulated as the contrastive learning \cite{Li:CPD20} between the visual and textual features. The features of the positive video-text pairs are pulled close while the features of negative pairs are pushed away. Following such setting, we experimented with contrastive learning on video and text via InfoNCE loss \cite{Oord:NIPS18} on the same backbone. The top-1 accuracy achieves 72.2\% on Kinetics-400 under linear evaluation protocol. The performance is lower than 72.5\% of the TS in Table \ref{table4:1}. Instead of learning the pair-wise correlation through contrastive learning, our TS mines the group-wise relationship via clustering. The results basically confirm that our TS is a good alternative for the video representation learning based on text information.

\begin{figure}[!tb]
	\centering
	\includegraphics[width=0.49\textwidth]{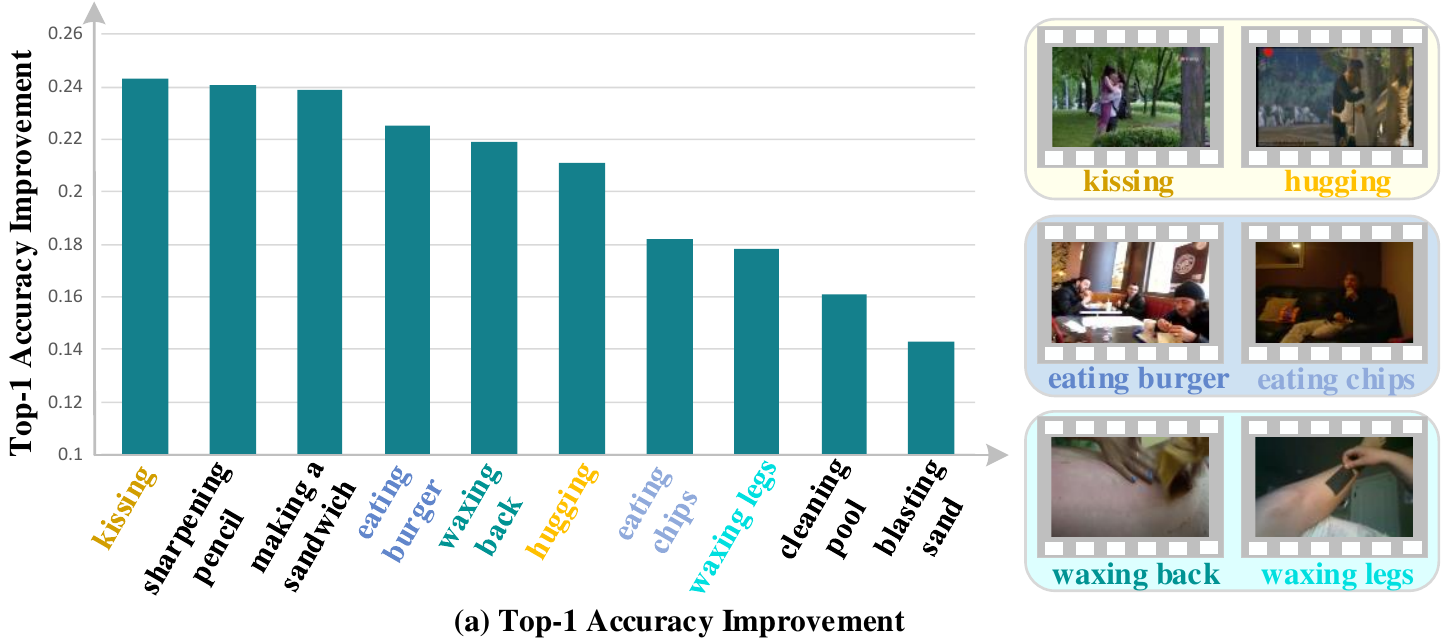}	\includegraphics[width=0.23\textwidth]{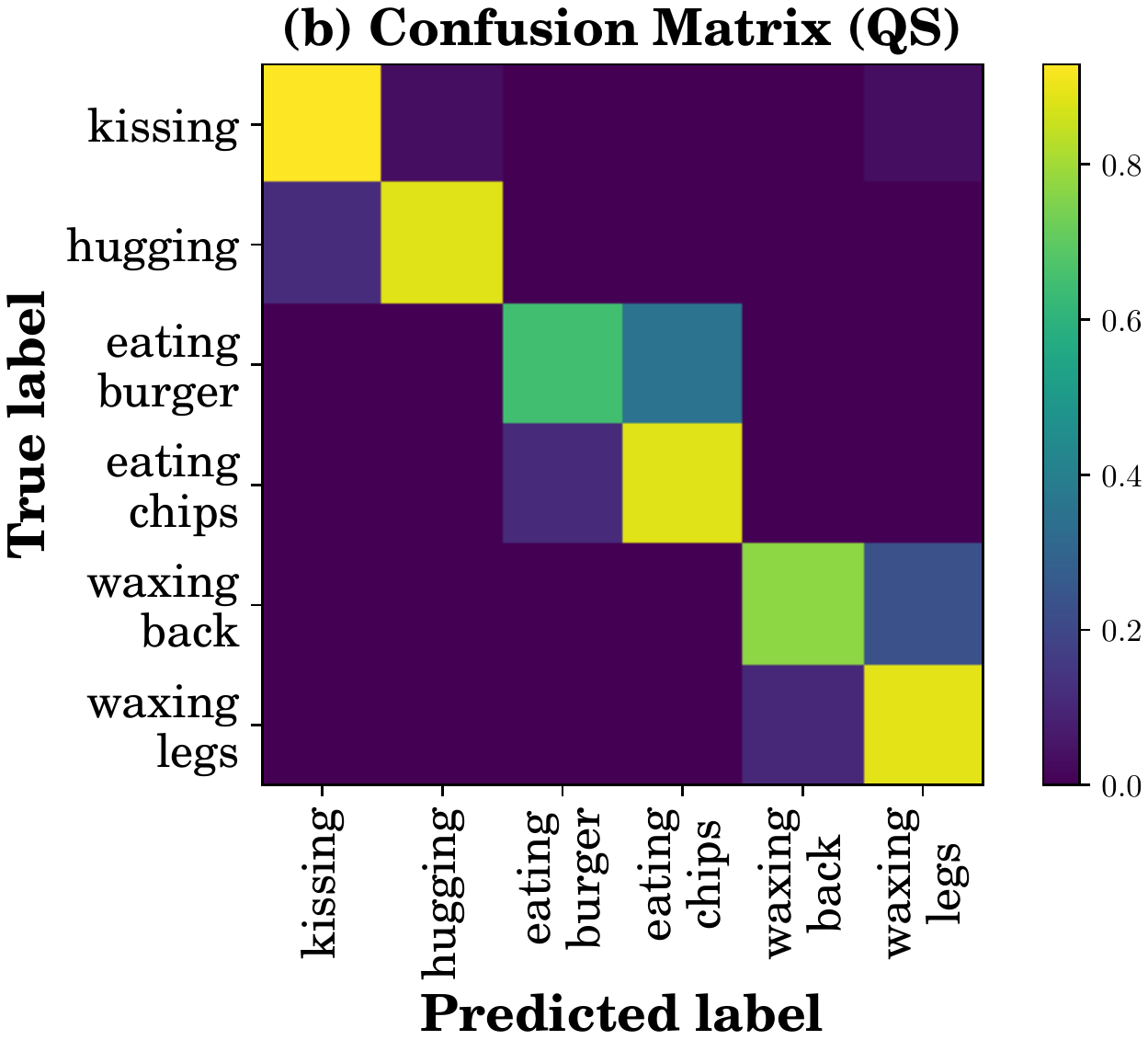} \hspace{0.05in}
	\includegraphics[width=0.23\textwidth]{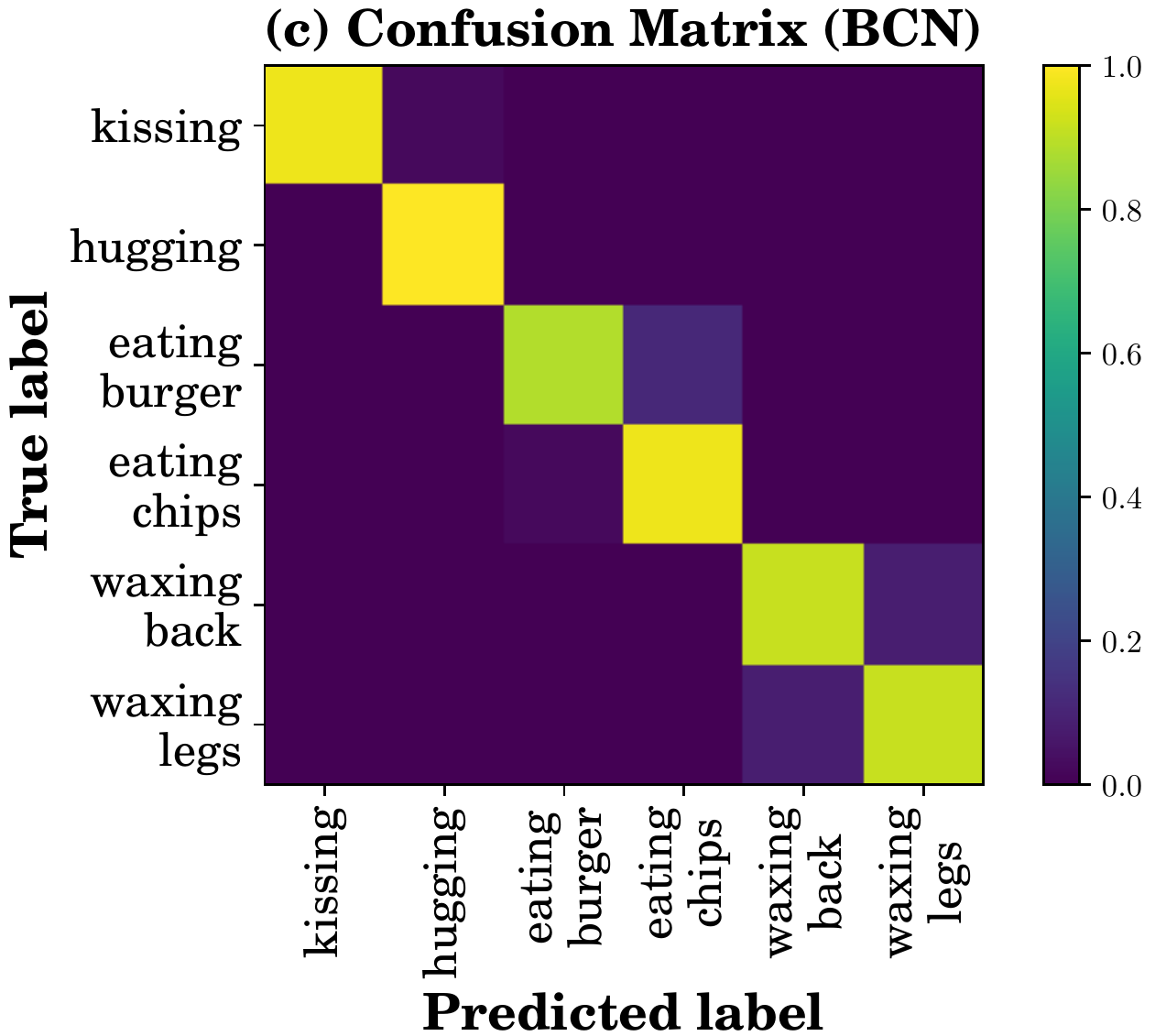}
	\caption{\small (a) 10 classes with the largest performance gain from QS to BCN under linear protocol on Kinetics-400 and the visualization of confusion matrix across the selected six categories on video representation learnt by (b) QS and (c) BCN.}
	\label{fig4:2}
	\vspace{-0.1in}
\end{figure}

\subsection{Evaluation on Bi-Calibration Networks}
Next, we study how each design in Bi-Calibration Networks influences video representation learning. QS and TS solely exploit primary query supervision and text supervision in our framework to guide video representation learning, respectively, through single-label and multi-label classification. QS+TS performs the joint learning on the primary supervision of query and text. BCN$_Q$/BCN$_T$ leverages the idea of text-to-query (T$\to$Q) or query-to-text (Q$\to$T) calibration to estimate t2q/q2t corrections to adjust primary query/text supervision and further boost video feature learning. BCN is our Bi-Calibration framework.

Table \ref{table4:2} details the top-1 accuracy on Kinetics-400 (K400), UCF101 (U101) and HMDB51 (HD51) datasets under linear model protocol by considering one more factor in our BCN. Compared to QS/TS, QS+TS boosts up the accuracy from 71.1\%/72.5\%, 90.7\%/91.8\% and 62.1\%/64.5\%, to 73.1\%, 92.3\% and 65.9\%, respectively, on three datasets. The results basically indicate that primary query supervision and primary text supervision are complementary to refine visual-textual connections and enhance video representation learning. Text-to-query/query-to-text calibration further amends primary query/text supervision and the performance gain of each is 0.7\%/0.5\%, 1.0\%/0.5\% and 1.0\%/0.4\% on three datasets against QS+TS. By mutual calibration between query and text, BCN finally reaches the top-1 accuracy of 74.2\%, 93.5\% and 67.6\%.

\begin{figure*}[!tb]
	\centering\includegraphics[width=0.99\textwidth]{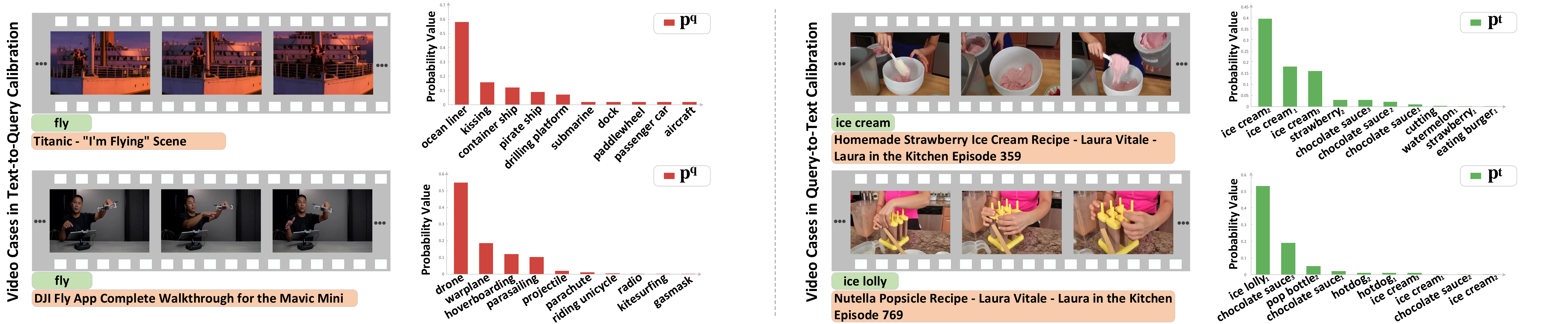}
	\caption{\small Video examples with the searched queries, video titles and the probability distribution on query/text ($\mathbf{p^q}$/$\mathbf{p^t}$).}
	\label{fig4:4}
	\vspace{-0.1in}
\end{figure*}

\begin{figure}[!tb]
	\subcaptionbox{}{\includegraphics[width=0.250\textwidth]{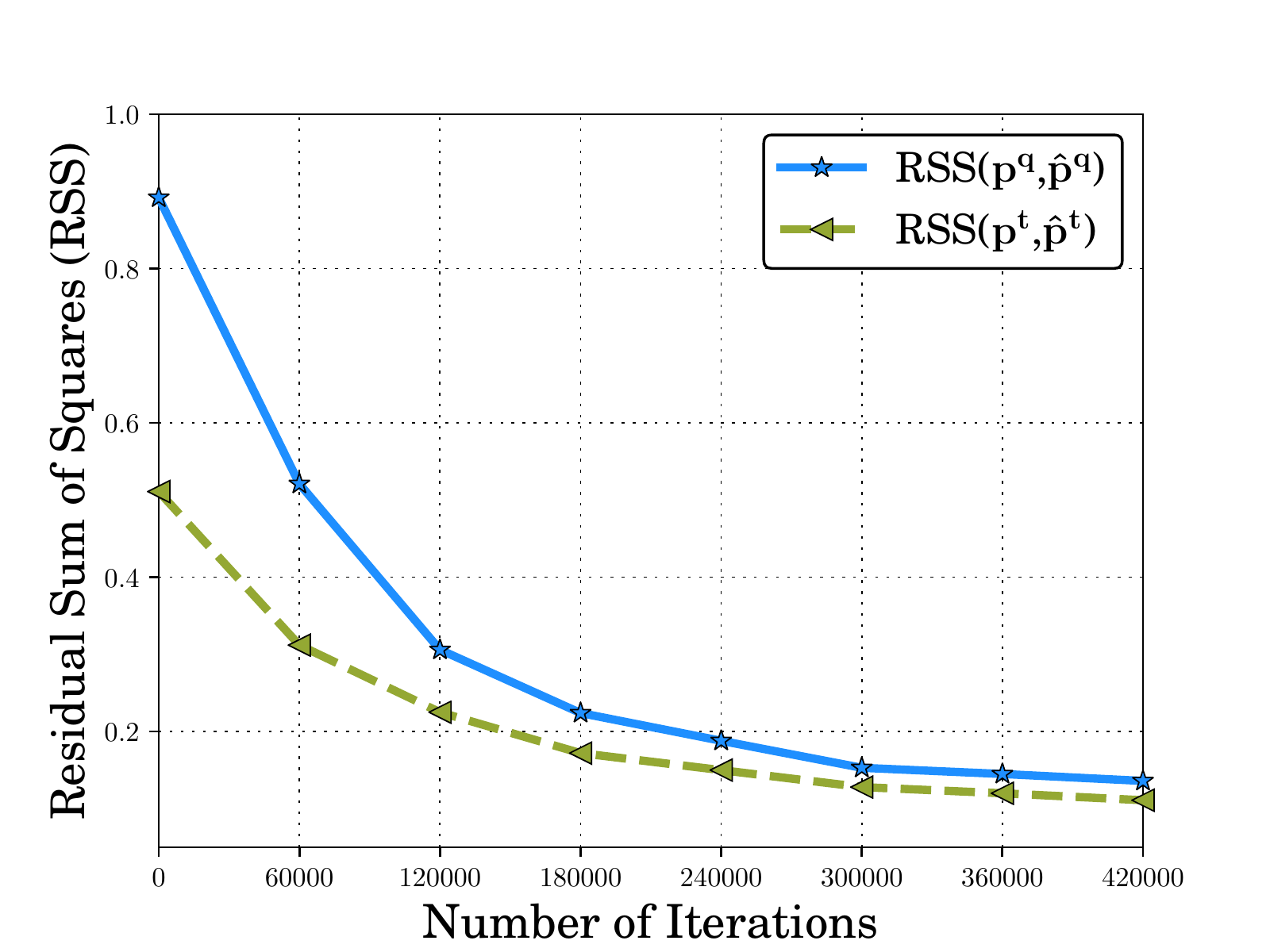}} \hspace{0.02in}
	\subcaptionbox{}{\includegraphics[width=0.223\textwidth]{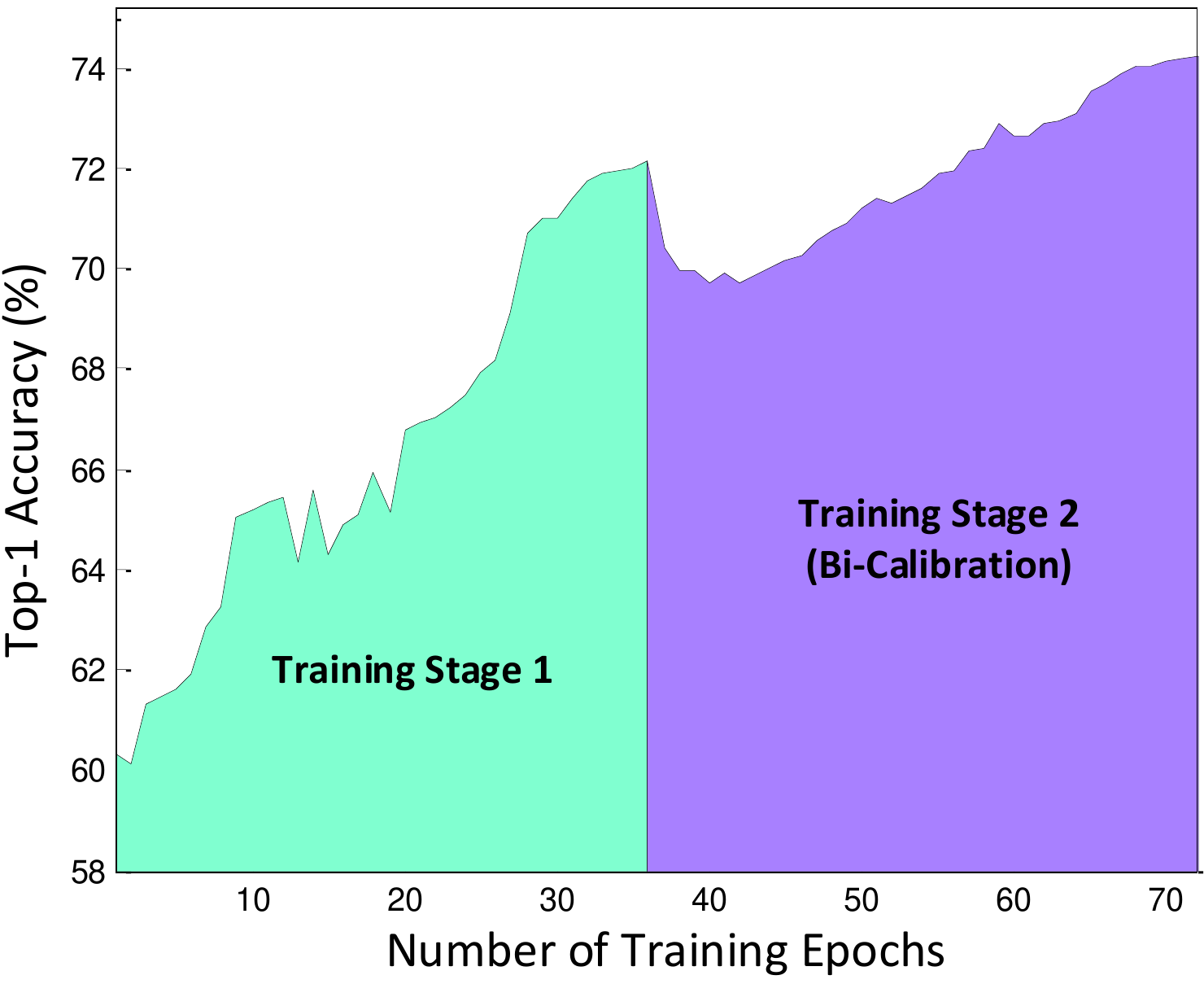}}
	\caption{\small (a) Residual Sum of Square (RSS) curves between $\mathbf{p^q}$/$\mathbf{p^t}$ and $\mathbf{\hat{p}^q}$/$\mathbf{\hat{p}^t}$ w.r.t the number of iterations. (b) Top-1 accuracy under linear protocol of two training stages on Kinetics-400. (The number of training epoch in each stage is 36 and decided through cross validation.)}
	\label{fig4:3}
\end{figure}

To verify the impact of mutual calibration via our BCN design across different categories, we further list the categories with most benefit. Figure \ref{fig4:2}(a) shows ten categories in Kinetics-400 which achieve the largest performance gain from QS to BCN. An interesting observation is that three pairs, i.e., kissing-hugging, eating burger-eating chips, waxing back-waxing legs, among the ten categories, are indeed fine-grained and it is challenging to distinguish one from the other in each pair. Figure \ref{fig4:2}(b) and (c) also visualizes confusion matrix across the six categories on video representation learnt by QS and BCN, respectively. It is clear that BCN endows video representation with more discriminative power especially on the fine-grained categories.

\subsection{More Analysis on Network Optimization}
The selection scheme controls the switch across the calibration of text-to-query or query-to-text directions. We compute the Residual Sum of Squares (RSS) between the probabilities ($\mathbf{p^q}$/$\mathbf{p^t}$) and t2q/q2t corrections ($\mathbf{\hat{p}^q}$/$\mathbf{\hat{p}^t}$), and Figure \ref{fig4:3}(a) depicts RSS curve with respect to the number of iterations. As expected, the RSS on both $(\mathbf{p^q}$,$\mathbf{\hat{p}^q})$ and $(\mathbf{p^t}$,$\mathbf{\hat{p}^t})$ is gradually decreased when training more iterations. The curve of RSS on $(\mathbf{p^t}$,$\mathbf{\hat{p}^t})$ changes more smoothly due to the momentum update strategy.
Since the corrections are adopted as the supervision to optimize the probabilities, the gradients are not back-propagated to them. 
Thus, the results give the clue that probabilities are enforced to be close to the corrections and BCN makes necessary corrections through text-to-query/query-to-text calibration, validating the impact of selection scheme. Figure \ref{fig4:3}(b) shows the curve of top-1 accuracy on Kinetics-400 under linear protocol in two optimization stages. Despite having some drops in accuracy at the beginning of bi-calibration training stage, the top-1 accuracy eventually improves and reaches a higher value, which again demonstrates the effectiveness of calibration between query and text through information refinement.

\begin{figure}[!tb]
	\centering\includegraphics[width=0.50\textwidth]{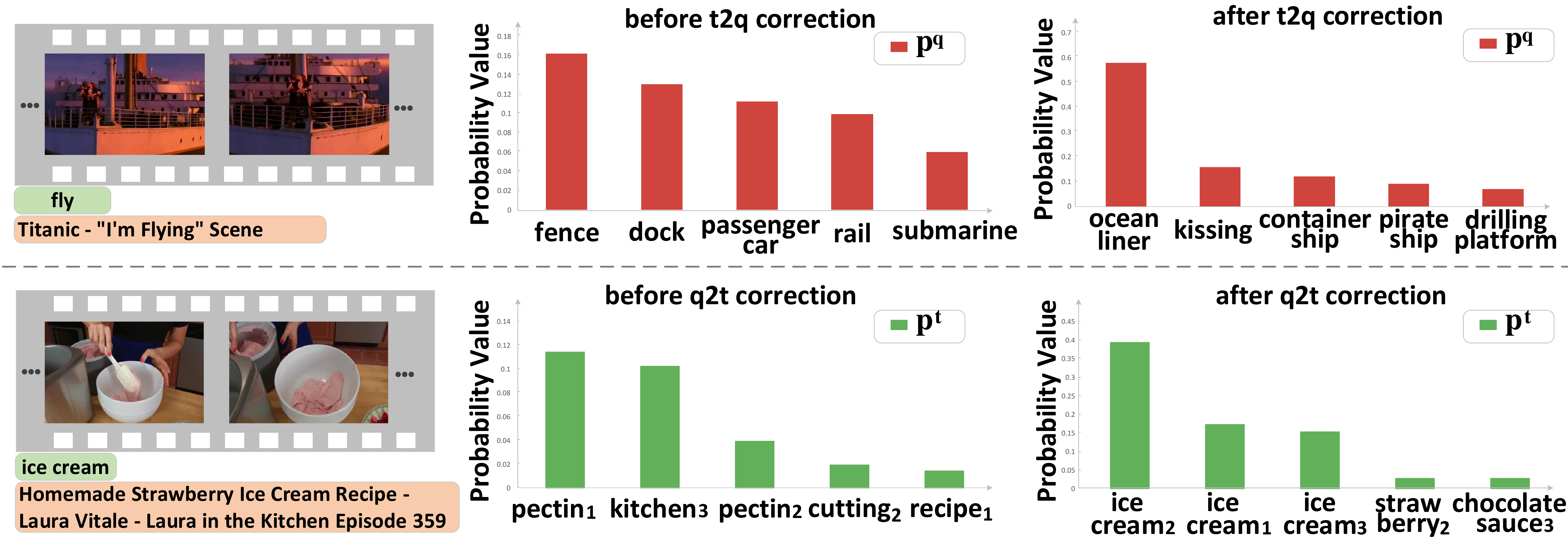}
	\caption{\small Probability distribution on query/text ($\mathbf{p^q}$/$\mathbf{p^t}$) of the two cases in Fig \ref{fig4:4} before and after the t2q/q2t correction.}
	\label{fig:prior}
\end{figure}

Figure \ref{fig4:4} showcases two groups of videos with the searched queries, video titles and the probabilities on query/text. The first group of videos are searched by an identical query of ``fly,'' but the video content corresponds to different meanings. One is about the famous scene on the ocean liner in movie ``Titanic'' and the other describes ``DJI'' drone. Such query polysemy may misdirect video representation learning. With the text-to-query corrections in BCN, the query probabilities of the two videos are well predicted. The video about ``Titanic'' is highly relevant to ``ocean liner'' and ``kissing,'' and the video of ``DJI'' has high probability in response to ``drone'' and ``warplane.'' In contrast, the second group of videos share similar syntactic structure of title, but are related to different queries of ``ice cream'' and ``ice lolly.'' In this case, solely capitalizing on title information may affect video representation learning as well. Through query-to-text correction, the text probabilities predicted by BCN nicely lead to different emphasis of ``ice cream'' or ``ice lolly'' and ``chocolate sauce.'' 
Figure \ref{fig:prior} further illustrates the probability distribution on query/text of two video cases in Figure \ref{fig4:4} before and after the t2q/q2t correction. For the ``Titanic'' video, the top-2 query prediction before t2q correction are ``fence'' and ``docker''. After the t2q calibration, we achieve more accurate predictions (``ocean liner'' and ``kissing'').   
Meanwhile, the q2t correction changes the original prediction of ``pectin'' to the preciser one of ``ice cream'' in the second video case.

\subsection{Comparisons with State-of-the-Art Methods}
We compare BCN with several state-of-the-art techniques on five datasets: Kinetics-400 (K400), UCF101 (U101) and HMDB51 (HD51) for scene-related action recognition, and Something-Something V1 (SS-V1) and V2 (SS-V2) for interaction-related action recognition.

\begin{table}[!tb]
	\setlength{\belowcaptionskip}{-1pt}
	\centering
	\caption{\small Performance comparisons on Kinetics-400.}
	\scalebox{0.90}[0.90]{
		\begin{tabular}{{l|c|c|c|c}}
			\hline
			\multicolumn{1}{c|}{\textbf{Approach}} & \textbf{Pre-training} & \textbf{Backbone} & \textbf{Top-1} & \textbf{Top-5}\\ \hline
			\multicolumn{5}{l}{\multirow{1}{*}{\textbf{Supervised Pre-training}}} \\ \hline
			\text{R(2+1)D RGB \cite{Tran:CVPR18}}                       &  ImageNet & custom    & 74.3 & 91.4 \\
			\text{I3D RGB \cite{Carreira:CVPR17}}                       &  ImageNet & Inception & 72.1 & 90.3 \\
			\text{S3D RGB \cite{Xie:ECCV18}}                            &  ImageNet & Inception & 74.7 & 93.4 \\
			\text{NL I3D RGB \cite{Wang:CVPR18}}                        &  ImageNet & ResNet-50 & 74.9 & 91.6 \\
			\text{TSM RGB \cite{JiLin:ICCV19}}                          &  ImageNet & ResNet-50 & 74.1 & 91.2 \\
			\text{LGD RGB \cite{Qiu:CVPR19}}                            &  ImageNet & ResNet-50 & 74.8 & 92.0 \\
			\text{SlowFast RGB \cite{Christoph:ICCV19}}                 &  ImageNet & ResNet-50 & 75.6 & 92.1 \\
			\text{SmallBig RGB \cite{Li:CVPR20}}                        &  ImageNet & ResNet-50 & 76.3 & 92.5 \\
			\text{TDN RGB \cite{Wang:CVPR21}}                           &  ImageNet & ResNet-50 & 77.5 & 93.2 \\
			\hline
			\multicolumn{5}{l}{\multirow{1}{*}{\textbf{Self-supervised Pre-training (linear protocol)}}} \\ \hline
			\text{VTHCL RGB \cite{Yang:VTHCL}}                          &  K400     & ResNet-50 & 37.8 &  -  \\
			\text{CVRL RGB \cite{Qian:CVPR21}}                          &  K400     & ResNet-50 & 66.1 &  -  \\
			$\rho$\text{BYOL RGB \cite{Feichtenhofer:CVPR21}}           &  K400     & ResNet-50 & 71.5 &  -  \\
			\hline
			\multicolumn{5}{l}{\multirow{1}{*}{\textbf{Weakly-supervised Pre-training}}} \\ \hline
			\multicolumn{5}{l}{\multirow{1}{*}{\text{Linear model protocol on video representation}}}           \\  \hline
			\text{CPD RGB\cite{Li:CPD20}}                                & K400-title& ResNet-50 & 63.8 &   -   \\
			\text{BCN RGB}                                              &  YOVO-3M    & ResNet-50 & 74.2 & 90.9  \\
			\text{BCN RGB}                                              &  YOVO-10M   & ResNet-50 & 74.9 & 91.6  \\ \hline
			\multicolumn{5}{l}{\multirow{1}{*}{\text{Network fine-tuning}}}                                            \\ \hline
			\text{CPD* RGB \cite{Li:CPD20}}                                 & YOVO-3M   & ResNet-50 & 73.9 &  90.2 \\
			\text{CPD* RGB \cite{Li:CPD20}}                                 & YOVO-10M  & ResNet-50 & 75.0 &  91.3 \\
			\text{BCN RGB}                                              & YOVO-3M   & ResNet-50 & 78.5 & 94.3 \\
			\text{BCN RGB}                                              & YOVO-10M  & ResNet-50 & \textbf{79.1}  &  \textbf{94.7}\\ \hhline{*{5}{-}}
		\end{tabular}
	}
	\label{table4:3}
\end{table}

Table \ref{table4:3} lists the top-1 and top-5 accuracy of different approaches on K400. For fair comparison, all the methods exploit only RGB modality of frame for model training and ``*'' denotes that the models are pre-trained on our YOVO-3M/10M with official source codes. Under linear model protocol, BCN achieves comparable performances with models built on fully-supervised ImageNet pre-training and self-supervised Kinetics pre-training.
When fine-tuning the pre-trained model by BCN on K400, BCN exhibits better performances against other baselines. In particular, BCN learnt on YOVO-3M obtains 78.5\% top-1 accuracy, which outperforms CPD \cite{Li:CPD20} by 4.6\% pre-trained on the same data.
Different from CPD which solely capitalizes on title information, BCN exploits mutual calibration between query and title to improve weakly-supervised video representation learning. Such result demonstrates the advantage of our bi-calibration design.
Compared to SmallBig and TDN, BCN leads the top-1 accuracy by 2.2\% and 1.0\%, respectively. Executing weakly-supervised learning on YOVO-10M further improves the accuracy from 78.5\% to 79.1\%.
The performance trends on U101 and HD51 are similar with those on K400 as shown in Table \ref{table4:4}. The results again verify the impact of BCN for weakly-supervised representation learning. Table \ref{table4:5} summarizes the performances on SS-V1 and SS-V2. Similarly, BCN under network fine-tuning protocol surpasses the best competitor SmallBig and ACTION-Net by 1.9\% and 1.8\% on SS-V1 and SS-V2, respectively.

\begin{table}[!tb]
	\setlength{\belowcaptionskip}{-1pt}
	\centering
	\caption{\small Top-1 accuracy on UCF101 and HMDB51.}
	\scalebox{0.85}[0.85]{
		\begin{tabular}{{l|c|c|c|c}}
			\hline
			\multicolumn{1}{c|}{\textbf{Approach}} & \textbf{Pre-training} & \textbf{Backbone}  & \textbf{U101} & \textbf{HD51}\\ \hline
			\multicolumn{5}{l}{\multirow{1}{*}{\textbf{Supervised Pre-training}}} \\ \hline
			\text{R(2+1)D RGB \cite{Tran:CVPR18}}                       &  K400     & custom    & 96.8 & 74.5     \\
			\text{I3D RGB \cite{Carreira:CVPR17}}                       &  Img+K400 & Inception & 95.4 & 74.5     \\
			\text{S3D RGB \cite{Xie:ECCV18}}                            &  Img+K400 & Inception & 96.8 & 75.9     \\
			\text{TSM RGB \cite{JiLin:ICCV19}}                          &  K400     & ResNet-50 & 95.9 & 73.5     \\
			\text{LGD RGB \cite{Qiu:CVPR19}}                            &  Img+K600 & ResNet-50 & 96.0 & 74.7     \\ \hline
			\multicolumn{5}{l}{\multirow{1}{*}{\textbf{Self-supervised Pre-training (fine-tuning)}}} \\ \hline
			\text{XDC RGB\cite{Alwassel:NIPS20}}                           &  K400      & R(2+1)D-18 & 84.2 & 47.1
			\\
			\text{SpeedNet RGB\cite{Benaim:CVPR20}}                        &  K400      & S3D-G       & 81.1 & 48.8
			\\
			\text{CoCLR RGB\cite{Han:NIPS20}}                              &  K400      & S3D-G       & 87.9 & 54.6
			\\
			\text{CVRL RGB\cite{Qian:CVPR21}}                              &  K400      & ResNet-50   & 92.2 & 66.7  \\
			$\rho$\text{BYOL RGB\cite{Feichtenhofer:CVPR21}}               &  K400      & ResNet-50   & 95.5 & 73.6  \\
			\hline
			\multicolumn{5}{l}{\multirow{1}{*}{\textbf{Weakly-supervised Pre-training}}} \\ \hline
			\multicolumn{5}{l}{\multirow{1}{*}{\text{Linear model protocol on video representation}}}                  \\ \hline
			\text{BCN RGB}                                              &  YOVO-3M  & ResNet-50 & 93.5 & 67.6     \\
			\text{BCN RGB}                                              &  YOVO-10M & ResNet-50 & 94.9 & 69.2     \\ \hline
			\multicolumn{5}{l}{\multirow{1}{*}{\text{Network fine-tuning}}}                                     \\ \hline
			\text{MIL-NCE RGB\cite{Miech:CVPR20}}                          &  HowTo100M & ResNet-50  & 91.3 & 61.0
			\\
			\text{CPD RGB\cite{Li:CPD20}}                                  & Instagram300k & ResNet-50 & 92.8 & 63.8   \\
			\text{CPD* RGB\cite{Li:CPD20}}                                 & YOVO-3M   & ResNet-50 & 94.2 & 70.2   \\
			\text{CPD* RGB\cite{Li:CPD20}}                                 & YOVO-10M  & ResNet-50 & 95.4 & 73.5   \\
			\text{BCN RGB}                                                 & YOVO-3M   & ResNet-50 & 97.1 & 74.9       \\
			\text{BCN RGB}                                                & YOVO-10M  & ResNet-50 &  \textbf{97.5} &  \textbf{76.5} \\ \hhline{*{5}{-}}
		\end{tabular}
	}
	\label{table4:4}
\end{table}

\begin{table}[!tb]
	\setlength{\belowcaptionskip}{-1pt}
	\centering
	\caption{\small Top-1 accuracy on Something-Something V1/V2.}
	\scalebox{0.78}[0.78]{
		\begin{tabular}{{l|c|c|c|c}}
			\hline
			\multicolumn{1}{c|}{\textbf{Approach}} & \textbf{Pre-training} & \textbf{Backbone} & \textbf{SS-V1} & \textbf{SS-V2} \\ \hline
			\multicolumn{5}{l}{\multirow{1}{*}{\textbf{Supervised Pre-training}}} \\ \hline
			\text{I3D RGB \cite{Wang:ECCV18}}                           &  ImageNet+K400 & ResNet-50 & 41.6 & -        \\
			\text{NL I3D RGB \cite{Wang:ECCV18}}                        &  ImageNet+K400 & ResNet-50 & 44.4 & -        \\
			\text{NL I3D + gcn RGB \cite{Wang:ECCV18}}                  &  ImageNet+K400 & ResNet-50 & 46.1 & -        \\
			\text{CPNet RGB \cite{Xingyu:CVPR19}}                       &  ImageNet      & ResNet-34 & -    & 57.7     \\
			\text{TSM RGB \cite{JiLin:ICCV19}}                          &  ImageNet      & ResNet-50 & 45.6 & 59.1     \\
			\text{SmallBig RGB \cite{Li:CVPR20}}                        &  ImageNet      & ResNet-50 & 48.3 & 61.6
			\\
			\text{ACTION-Net RGB \cite{WangAct:CVPR21}}                 &  ImageNet      & ResNet-50 &  -   & 62.5
			\\ \hline
			\multicolumn{5}{l}{\multirow{1}{*}{\textbf{Self-supervised Pre-training (fine-tuning)}}} \\ \hline
			\text{BYOL RGB\cite{Feichtenhofer:CVPR21}}              &  K400      & ResNet-50   &  -   & 55.8
			\\
			\text{MoCo RGB\cite{Feichtenhofer:CVPR21}}              &  K400      & ResNet-50   &  -   & 54.4
			\\ \hline
			\multicolumn{5}{l}{\multirow{1}{*}{\textbf{Weakly-supervised Pre-training}}} \\ \hline
			\multicolumn{5}{l}{\multirow{1}{*}{\text{Linear model protocol on video representation}}}           \\ \hline
			\text{BCN RGB}                                          &  YOVO-3M       & ResNet-50 & 42.7 & 53.2     \\
			\text{BCN RGB}                                          &  YOVO-10M      & ResNet-50 & 43.2 & 55.7     \\ \hline
			\multicolumn{5}{l}{\multirow{1}{*}{\text{Network fine-tuning}}}                                     \\ \hline
			\text{CPD* RGB\cite{Li:CPD20}}                          & YOVO-3M   & ResNet-50 & 45.2 & 60.1   \\
			\text{CPD* RGB\cite{Li:CPD20}}                          & YOVO-10M  & ResNet-50 & 47.1 & 61.7   \\
			\text{BCN RGB}                                          & YOVO-3M        & ResNet-50 & 48.6 & 62.6       \\
			\text{BCN RGB}                                          & YOVO-10M       & ResNet-50 &  \textbf{50.2} &  \textbf{64.3} \\ \hhline{*{5}{-}}
		\end{tabular}
	}
	\label{table4:5}
\end{table}

\section{Conclusions and Discussions}
We have presented Bi-Calibration Networks (BCN), which explores the correlations between web videos and the searched queries or video titles for improving weakly-supervised video representation learning. Particularly, we study the problem from the viewpoint of refining the visual-semantic connections through mutual calibration between query and title information. To materialize our idea, we first achieve the primary query and text supervision on query vocabulary of query words and text vocabulary of text prototypes, which are utilized to optimize video-to-query (v2q) and video-to-text (v2t) projections for classification. Next, the v2t/v2q projection triggers the text-to-query or query-to-text calibration, that aims to adjust primary query/text supervision to further optimize v2q/v2t projection. Extensive experiments conducted on newly-created web video datasets, i.e., YOVO-3M and YOVO-10M, validate our BCN. More remarkably, weakly-supervised pre-training BCN on YOVO-10M is superior to several techniques with fully-supervised ImageNet or Kinetics pre-training.


%

%
%
%
%
%
%
%
%
%
%
%

\bibliographystyle{IEEEtran}
\bibliography{egbib}

\begin{thebibliography}{10}
\providecommand{\url}[1]{#1}
\csname url@samestyle\endcsname
\providecommand{\newblock}{\relax}
\providecommand{\bibinfo}[2]{#2}
\providecommand{\BIBentrySTDinterwordspacing}{\spaceskip=0pt\relax}
\providecommand{\BIBentryALTinterwordstretchfactor}{4}
\providecommand{\BIBentryALTinterwordspacing}{\spaceskip=\fontdimen2\font plus
\BIBentryALTinterwordstretchfactor\fontdimen3\font minus
  \fontdimen4\font\relax}
\providecommand{\BIBforeignlanguage}[2]{{%
\expandafter\ifx\csname l@#1\endcsname\relax
\typeout{** WARNING: IEEEtran.bst: No hyphenation pattern has been}%
\typeout{** loaded for the language `#1'. Using the pattern for}%
\typeout{** the default language instead.}%
\else
\language=\csname l@#1\endcsname
\fi
#2}}
\providecommand{\BIBdecl}{\relax}
\BIBdecl

\bibitem{Carreira:CVPR17}
J.~Carreira and A.~Zisserman, ``{Quo Vadis, Action Recognition? A New Model and
  the Kinetics Dataset},'' in \emph{CVPR}, 2017.

\bibitem{Yan:CVPR20}
Y.~Li, B.~Ji, X.~Shi, J.~Zhang, B.~Kang, and L.~Wang, ``{TEA: Temporal
  Excitation and Aggregation for Action Recognition},'' in \emph{CVPR}, 2020.

\bibitem{Simonyan:NIPS14}
K.~Simonyan and A.~Zisserman, ``{Two-stream Convolutional Networks for Action
  Recognition in Videos},'' in \emph{NIPS}, 2014.

\bibitem{Yang:CVPR20}
C.~Yang, Y.~Xu, J.~Shi, B.~Dai, and B.~Zhou, ``{Temporal Pyramid Network for
  Action Recognition},'' in \emph{CVPR}, 2020.

\bibitem{Liu:V-Swin}
Z.~Liu, J.~Ning, Y.~Cao, Y.~Wei, Z.~Zhang, S.~Lin, and H.~Hu, ``{Video Swin
  Transformer},'' \emph{arXiv preprint arXiv:2106.13230}, 2021.

\bibitem{C3D:PAMI}
S.~Ji, W.~Xu, M.~Yang, and K.~Yu, ``{3D Convolutional Neural Networks for Human
  Action Recognition},'' \emph{IEEE Trans. on PAMI}, vol.~35, no.~1, pp.
  221--231, 2012.

\bibitem{T1D:PAMI}
G.~Varol, I.~Laptev, and C.~Schmid, ``{Long-Term Temporal Convolutions for
  Action Recognition},'' \emph{IEEE Trans. on PAMI}, vol.~40, no.~6, pp.
  1510--1517, 2017.

\bibitem{Wang:PAMI20}
X.~Wang, L.~Zhu, Y.~Wu, and Y.~Yang, ``{Symbiotic Attention for Egocentric
  Action Recognition with Object-centric Alignment},'' \emph{IEEE Trans. on
  PAMI}, vol.~1, no.~1, pp. 1--13, 2020.

\bibitem{Gaidon:PAMI13}
A.~Gaidon, Z.~Harchaoui, and C.~Schmid, ``{Temporal Localization of Actions
  with Actoms},'' \emph{IEEE Trans. on PAMI}, vol.~35, no.~11, pp. 2782--2795,
  2013.

\bibitem{Ghadiyaram:CVPR19}
D.~Ghadiyaram, M.~Feiszli, D.~Tran, X.~Yan, H.~Wang, and D.~Mahajan,
  ``{Large-scale weakly-supervised pre-training for video action
  recognition},'' in \emph{CVPR}, 2019.

\bibitem{Li:CPD20}
T.~Li and L.~Wang, ``{Learning Spatiotemporal Features via Video and Text Pair
  Discrimination},'' \emph{arXiv preprint arXiv:2001.05691}, 2020.

\bibitem{Jacob:ACL18}
J.~Devlin, M.-W. Chang, K.~Lee, and K.~Toutanova, ``{BERT: Pre-training of Deep
  Bidirectional Transformers for Language Understanding},'' in \emph{ACL},
  2018.

\bibitem{Diba:CVPR17}
A.~Diba, V.~Sharma, and L.~V. Gool, ``{Deep Temporal Linear Encoding
  Networks},'' in \emph{CVPR}, 2017.

\bibitem{Feichtenhofer:CVPR16}
C.~Feichtenhofer, A.~Pinz, and A.~Zisserman, ``{Convolutional Two-Stream
  Network Fusion for Video Action Recognition},'' in \emph{CVPR}, 2016.

\bibitem{Sports1M}
A.~Karpathy, G.~Toderici, S.~Shetty, T.~Leung, R.~Sukthankar, and L.~Fei-Fei,
  ``{Large-scale Video Classification with Convolutional Neural Networks},'' in
  \emph{CVPR}, 2014.

\bibitem{Yue-Hei:CVPR15}
J.~Y.-H. Ng, M.~Hausknecht, S.~Vijayanarasimhan, O.~Vinyals, R.~Monga, and
  G.~Toderici, ``{Beyond Short Snippets: Deep Networks for Video
  Classification},'' in \emph{CVPR}, 2015.

\bibitem{Wang:ECCV16}
L.~Wang, Y.~Xiong, Z.~Wang, Y.~Qiao, D.~Lin, X.~Tang, and L.~V. Gool,
  ``{Temporal Segment Networks: Towards Good Practices for Deep Action
  Recognition},'' in \emph{ECCV}, 2016.

\bibitem{TSN:PAMI}
L.~Wang, Y.~Xiong, Z.~Wang, Y.~Qiao, D.~Lin, X.~Tang, and L.~Gool, ``{Temporal
  Segment Networks for Action Recognition in Videos},'' \emph{IEEE Trans. on
  PAMI}, vol.~41, no.~11, pp. 2740--2755, 2018.

\bibitem{Hara:CVPR18}
K.~Hara, H.~Kataoka, and Y.~Satoh, ``{Can Spatiotemporal 3D CNNs Retrace the
  History of 2D CNNs and ImageNet?}'' in \emph{CVPR}, 2018.

\bibitem{Qiu:ICCV17}
Z.~Qiu, T.~Yao, and T.~Mei, ``{Learning Spatio-Temporal Representation with
  Pseudo-3D Residual Networks},'' in \emph{ICCV}, 2017.

\bibitem{Tran:ICCV15}
D.~Tran, L.~Bourdev, R.~Fergus, L.~Torresani, and M.~Paluri, ``{Learning
  Spatiotemporal Features with 3D Convolutional Networks},'' in \emph{ICCV},
  2015.

\bibitem{Tran:CVPR18}
D.~Tran, H.~Wang, L.~Torresani, J.~Ray, Y.~LeCun, and M.~Paluri, ``{A Closer
  Look at Spatiotemporal Convolutions for Action Recognition},'' in
  \emph{CVPR}, 2018.

\bibitem{Kinetics:600}
B.~Ghanem, J.~C. Niebles, C.~Snoek, F.~C. Heilbron, H.~Alwassel, V.~Escorcia,
  R.~Krishna, S.~Buch, and C.~D. Dao, ``{The ActivityNet Large-Scale Activity
  Recognition Challenge 2018 Summary},'' \emph{arXiv preprint
  arXiv:1808.03766}, 2018.

\bibitem{Bertasius:ICML21}
G.~Bertasius, H.~Wang, and L.~Torresani, ``{Is Space-Time Attention All You
  Need for Video Understanding?}'' in \emph{ICML}, 2021.

\bibitem{Fan:MVIT}
H.~Fan, B.~Xiong, K.~Mangalam, Y.~Li, Z.~Yan, J.~Malik, and C.~Feichtenhofer,
  ``{Multiscale Vision Transformers},'' \emph{arXiv preprint arXiv:2104.11227},
  2021.

\bibitem{ViViT}
A.~Arnab, M.~Dehghani, G.~Heigold, C.~Sun, M.~Lucic, and C.~Schmid, ``{ViViT: A
  Video Vision Transformer},'' in \emph{ICCV}, 2021.

\bibitem{ViT}
A.~Dosovitskiy, L.~Beyer, A.~Kolesnikov, D.~Weissenborn, X.~Zhai,
  T.~Unterthiner, M.~Dehghani, M.~Minderer, G.~Heigold, S.~Gelly, J.~Uszkoreit,
  and N.~Houlsby, ``{An Image is Worth 16x16 Words: Transformers for Image
  Recognition at Scale},'' in \emph{ICLR}, 2021.

\bibitem{Swin-ViT}
Z.~Liu, Y.~Lin, Y.~Cao, H.~Hu, Y.~Wei, Z.~Zhang, S.~Lin, and B.~Guo, ``{Swin
  Transformer: Hierarchical Vision Transformer using Shifted Windows},'' in
  \emph{ICCV}, 2021.

\bibitem{Fernando:CVPR17}
B.~Fernando, H.~Bilen, E.~Gavves, and S.~Gould, ``{Self-Supervised Video
  Representation Learning With Odd-One-Out Networks},'' in \emph{CVPR}, 2017.

\bibitem{Misra:ECCV16}
I.~Misra, C.~L. Zitnick, and M.~Hebert, ``{Shuffle and Learn: Unsupervised
  Learning using Temporal Order Verification},'' in \emph{ECCV}, 2016.

\bibitem{Wei:CVPR18}
D.~Wei, J.~Lim, A.~Zisserman, and W.~T. Freeman, ``{Learning and Using the
  Arrow of Time},'' in \emph{CVPR}, 2018.

\bibitem{Xu:CVPR19}
D.~Xu, J.~Xiao, Z.~Zhao, J.~Shao, D.~Xie, and Y.~Zhuang, ``{Self-supervised
  Spatiotemporal Learning via Video Clip Order Prediction},'' in \emph{CVPR},
  2019.

\bibitem{Agrawal:ICCV15}
P.~Agrawal, J.~Carreira, and J.~Malik, ``{Learning to See by Moving},'' in
  \emph{ICCV}, 2015.

\bibitem{Pathak:CVPR17}
D.~Pathak, R.~Girshick, P.~Dollar, T.~Darrell, and B.~Hariharan, ``{Learning
  Features by Watching Objects Move},'' in \emph{CVPR}, 2017.

\bibitem{Dwibedi:CVPR19}
D.~Dwibedi, Y.~Aytar, J.~Tompson, P.~Sermanet, and A.~Zisserman, ``{Temporal
  Cycle-Consistency Learning},'' in \emph{CVPR}, 2019.

\bibitem{Wang:CVPR19}
X.~Wang, A.~Jabri, and A.~A. Efros, ``{Learning Correspondence from the
  Cycle-Consistency of Time},'' in \emph{CVPR}, 2019.

\bibitem{Mobahi:ICML09}
H.~Mobahi, R.~Collobert, and J.~Weston, ``{Deep Learning from Temporal
  Coherence in Video},'' in \emph{ICML}, 2009.

\bibitem{Wang:ICCV15}
X.~Wang and A.~Gupta, ``{Unsupervised Learning of Visual Representations using
  Videos},'' in \emph{ICCV}, 2015.

\bibitem{Liu:ICCV17}
Z.~Liu, R.~A. Yeh, X.~Tang, Y.~Liu, and A.~Agarwala, ``{Video Frame Synthesis
  using Deep Voxel Flow},'' in \emph{ICCV}, 2017.

\bibitem{Finn:NIPS16}
C.~Finn, I.~Goodfellow, and S.~Levine, ``{Unsupervised Learning for Physical
  Interaction through Video Prediction},'' in \emph{NIPS}, 2016.

\bibitem{Luo:CVPR17}
Z.~Luo, B.~Peng, D.-A. Huang, A.~Alahi, and L.~Fei-Fei, ``{Unsupervised
  Learning of Long-Term Motion Dynamics for Videos},'' in \emph{CVPR}, 2017.

\bibitem{Srivastava:ICML15}
N.~Srivastava, E.~Mansimov, and R.~Salakhutdinov, ``{Unsupervised Learning of
  Video Representations using LSTMs},'' in \emph{ICML}, 2015.

\bibitem{Feichtenhofer:CVPR21}
C.~Feichtenhofer, H.~Fan, B.~Xiong, R.~Girshick, and K.~He, ``{A Large-Scale
  Study on Unsupervised Spatiotemporal Representation Learning},'' in
  \emph{CVPR}, 2021.

\bibitem{Li:ICLR21}
J.~Li, P.~Zhou, C.~Xiong, and S.~Hoi, ``{Prototypical Contrastive Learning of
  Unsupervised Representations},'' in \emph{ICLR}, 2021.

\bibitem{Miech:CVPR20}
A.~Miech, J.-B. Alayrac, L.~Smaira, I.~Laptev, J.~Sivic, and A.~Zisserman,
  ``{End-to-End Learning of Visual Representations from Uncurated Instructional
  Videos},'' in \emph{CVPR}, 2020.

\bibitem{Miech:ICCV19}
A.~Miech, D.~Zhukov, J.-B. Alayrac, M.~Tapaswi, I.~Laptev, and J.~Sivic,
  ``{HowTo100M: Learning a Text-Video Embedding by Watching Hundred Million
  Narrated Video Clips},'' in \emph{ICCV}, 2019.

\bibitem{Berg:CVPR06}
T.~L. Berg and D.~A. Forsyth, ``{Animals on Web},'' in \emph{CVPR}, 2006.

\bibitem{Saenko:NIPS08}
K.~Saenko and T.~Darrell, ``{Unsupervised Learning of Visual Sense Models for
  Polysemous Words},'' in \emph{NIPS}, 2008.

\bibitem{Schroff:ICCV07}
F.~Schroff, A.~Criminisi, and A.~Zisserman, ``{Harvesting Image Databases from
  the Web},'' in \emph{ICCV}, 2007.

\bibitem{Stroud:Trp20}
J.~C. Stroud, D.~A. Ross, C.~Sun, J.~Deng, R.~Sukthankar, and C.~Schmid,
  ``{Learning Video Representations from Textual Web Supervision},''
  \emph{arXiv preprint arXiv:2007.14937}, 2020.

\bibitem{Qiu:CVPR21}
Z.~Qiu, T.~Yao, C.-W. Ngo, X.-P. Zhang, D.~Wu, and T.~Mei, ``{Boosting Video
  Representation Learning with Multi-Faceted Integration},'' in \emph{CVPR},
  2021.

\bibitem{Maaten:JMLR8}
L.~van~der Maaten and G.~Hinton, ``{Visualizing Data using t-SNE},''
  \emph{JMLR}, 2008.

\bibitem{Lazebnik:CVPR06}
S.~Lazebnik, C.~Schmid, and J.~Ponce, ``{Beyond Bags of Features: Spatial
  Pyramid Matching for Recognizing Natural Scene Categories},'' in \emph{CVPR},
  2006.

\bibitem{Tibshirani:Royal01}
R.~Tibshirani, G.~Walther, and T.~Hastie, ``{Estimating the number of cluster
  in a data set via the gap statistic},'' \emph{Journal of the Royal
  Statistical Society: Series B}, pp. 411--423, 2001.

\bibitem{Soomro:UCF101}
K.~Soomro, A.~R. Zamir, and M.~Shah, ``{UCF101: A Dataset of 101 Human Actions
  Classes From Videos in The Wild},'' \emph{CRCV-TR-12-01}, 2012.

\bibitem{Kuehne:ICCV11}
H.~Kuehne, H.~Jhuang, E.~Garrote, T.~Poggio, and T.~Serre, ``{HMDB: A Large
  Video Database for Human Motion Recognition},'' in \emph{ICCV}, 2011.

\bibitem{Goyal:SS}
R.~Goyal, S.~E. Kahou, V.~Michalski, J.~Materzynska, S.~Westphal, H.~Kim,
  V.~Haenel, I.~Fruend, P.~Yianilos, M.~Mueller-Freitag, F.~Hoppe, C.~Thurau,
  I.~Bax, and R.~Memisevic, ``{The "something something" video database for
  learning and evaluating visual common sense},'' in \emph{ICCV}, 2017.

\bibitem{Kinetics700}
J.~Carreira, E.~Noland, C.~Hillier, and A.~Zisserman, ``{A Short Note on the
  Kinetics-700 Human Action Dataset},'' \emph{arXiv preprint arXiv:1907.06987},
  2019.

\bibitem{ImageNetDatabase}
J.~Deng, W.~Dong, R.~Socher, L.-J. Li, K.~Li, and L.~Fei-Fei, ``{ImageNet: A
  large-scale hierarchical image database},'' in \emph{CVPR}, 2009.

\bibitem{Moments}
M.~Monfort, A.~Andonian, B.~Zhou, K.~Ramakrishnan, S.~A. Bargal, T.~Yan,
  L.~Brown, Q.~Fan, D.~Gutfreund, C.~Vondrick, and A.~Oliva, ``{Moments in Time
  Dataset: One Million Videos for Event Understanding},'' \emph{IEEE Trans. on
  PAMI}, vol.~42, no.~2, pp. 502--508, 2019.

\bibitem{Zasbih:ECCV94}
R.~Zabih and J.~Woodfill, ``{Non-parametric Local Transforms for Computing
  Visual Correspondence},'' in \emph{ECCV}, 1994.

\bibitem{Gionis:VLDB99}
A.~Gionis, P.~Indyk, and R.~Motwani, ``{Similarity Search in High Dimensions
  via Hashing},'' in \emph{VLDB}, 1999.

\bibitem{Qiu:CVPR19}
Z.~Qiu, T.~Yao, C.-W. Ngo, X.~Tian, and T.~Mei, ``{Learning Spatio-Temporal
  Representation with Local and Global Diffusion},'' in \emph{CVPR}, 2019.

\bibitem{Kaiming:CVPR16}
K.~He, X.~Zhang, S.~Ren, and J.~Sun, ``{Deep Residual Learning for Image
  Recognition},'' in \emph{CVPR}, 2016.

\bibitem{Yang:caffe}
Y.~Jia, E.~Shelhamer, J.~Donahue, S.~Karayev, J.~Long, R.~B. Girshick,
  S.~Guadarrama, and T.~Darrell, ``{Caffe: Convolutional Architecture for Fast
  Feature Embedding},'' in \emph{ACM MM}, 2014.

\bibitem{Girshick:ICCV15}
R.~Girshick, ``{Fast R-CNN},'' in \emph{ICCV}, 2015.

\bibitem{He:CVPR20}
K.~He, H.~Fan, Y.~Wu, S.~Xie, and R.~Girshick, ``{Momentum Contrast for
  Unsupervised Visual Representation Learning},'' in \emph{CVPR}, 2020.

\bibitem{Oord:NIPS18}
A.~van~den Oord, Y.~Li, and O.~Vinyals, ``{Representation Learning with
  Contrastive Predictive Coding},'' in \emph{NIPS}, 2018.

\bibitem{Xie:ECCV18}
S.~Xie, C.~Sun, J.~Huang, Z.~Tu, and K.~Murphy, ``{Rethinking Spatiotemporal
  Feature Learning: Speed-Accuracy Trade-offs in Video Classification},'' in
  \emph{ECCV}, 2018.

\bibitem{Wang:CVPR18}
X.~Wang, R.~Girshick, A.~Gupta, and K.~He, ``{Non-local Neural Networks},'' in
  \emph{CVPR}, 2018.

\bibitem{JiLin:ICCV19}
J.~Lin, C.~Gan, and S.~Han, ``{TSM: Temporal Shift Module for Efficient Video
  Understanding},'' in \emph{ICCV}, 2019.

\bibitem{Christoph:ICCV19}
C.~Feichtenhofer, H.~Fan, J.~Malik, and K.~He, ``{SlowFast Networks for Video
  Recognition},'' in \emph{ICCV}, 2019.

\bibitem{Li:CVPR20}
X.~Li, Y.~Wang, Z.~Zhou, and Y.~Qiao, ``{SmallBigNet: Integrating Core and
  Contextual Views for Video Classification},'' in \emph{CVPR}, 2020.

\bibitem{Wang:CVPR21}
L.~Wang, Z.~Tong, B.~Ji, and G.~Wu, ``{TDN: Temporal Difference Networks for
  Efficient Action Recognition},'' in \emph{CVPR}, 2021.

\bibitem{Yang:VTHCL}
C.~Yang, Y.~Xu, B.~Dai, and B.~Zhou, ``{Video Representation Learning with
  Visual Tempo Consistency},'' \emph{arXiv preprint arXiv:2006.15489}, 2020.

\bibitem{Qian:CVPR21}
R.~Qian, T.~Meng, B.~Gong, M.-H. Yang, H.~Wang, S.~Belongie, and Y.~Cui,
  ``{Spatiotemporal Contrastive Video Representation Learning},'' in
  \emph{CVPR}, 2021.

\bibitem{Alwassel:NIPS20}
H.~Alwassel, D.~Mahajan, B.~Korbar, L.~Torresani, B.~Ghanem, and D.~Tran,
  ``{Self-Supervised Learning by Cross-Modal Audio-Video Clustering},'' in
  \emph{NeurIPS}, 2020.

\bibitem{Benaim:CVPR20}
S.~Benaim, A.~Ephrat, O.~Lang, I.~Mosseri, W.~T. Freeman, M.~Rubinstein,
  M.~Irani, and T.~Dekel, ``{SpeedNet: Learning the Speediness in Videos},'' in
  \emph{CVPR}, 2020.

\bibitem{Han:NIPS20}
T.~Han, W.~Xie, and A.~Zisserman, ``{Self-supervised Co-training for Video
  Representation Learning},'' in \emph{NeurIPS}, 2020.

\bibitem{Wang:ECCV18}
X.~Wang and A.~Gupta, ``{Videos as Space-Time Region Graphs},'' in \emph{ECCV},
  2018.

\bibitem{Xingyu:CVPR19}
X.~Liu, J.-Y. Lee, and H.~Jin, ``{Learning Video Representations from
  Correspondence Proposals},'' in \emph{CVPR}, 2019.

\bibitem{WangAct:CVPR21}
Z.~Wang, Q.~She, and A.~Smolic, ``{ACTION-Net: Multipath Excitation for Action
  Recognition},'' in \emph{CVPR}, 2021.

\end{thebibliography}

\begin{IEEEbiography}[{\includegraphics[width=1in,height=1.25in,clip]{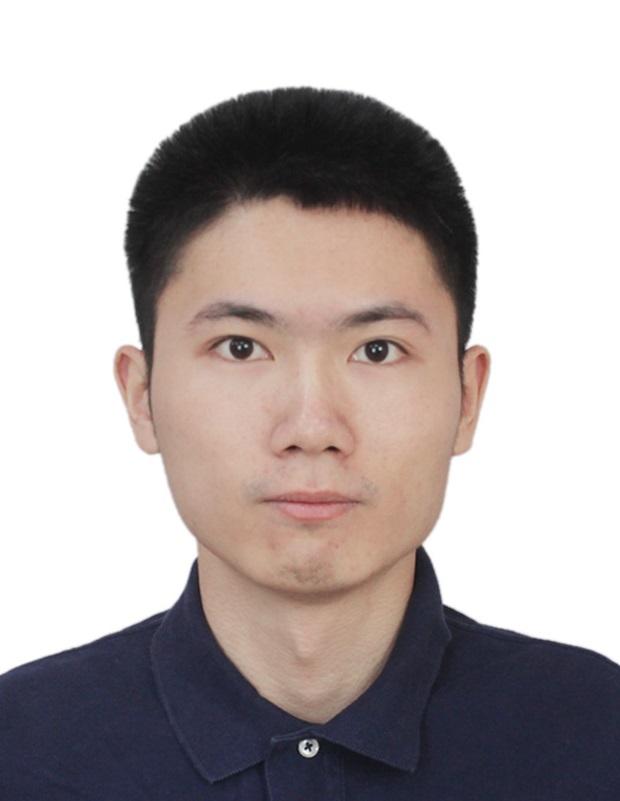}}]{Fuchen Long}
 is currently a Researcher in Vision and Multimedia Lab at JD Explore Academy, Beijing, China. He has participated several temporal action proposal and detection competitions such as Activity detection in Extended Videos (ActEV-PC) in ActivityNet Challenge 2019, ActivityNet Temporal Action Detection Challenge 2018 and ActivityNet Temporal Action Proposal Challenge 2017. His research interests include temporal action proposal and localization, multimedia retrieval and video understanding. He received Ph.D. degree from the University of Science and Technology of China (USTC) in 2021.
\end{IEEEbiography}

\begin{IEEEbiography}[{\includegraphics[width=1in,height=1.25in,clip]{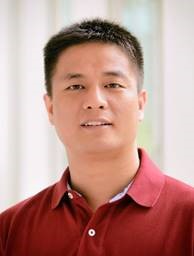}}]{Ting Yao}
	is currently a Principal Researcher in Vision and Multimedia Lab at JD Explore Academy, Beijing, China. His research interests include video understanding, vision and language, and deep learning. Prior to joining JD.com, he was a Researcher with Microsoft Research Asia, Beijing, China. Ting is the principal designer of several top-performing multimedia analytic systems in international benchmark competitions such as ActivityNet Large Scale Activity Recognition Challenge 2019-2016, Visual Domain Adaptation Challenge 2019-2017, and COCO Image Captioning Challenge. He is the leader organizer of MSR Video to Language Challenge in ACM Multimedia 2017 \& 2016, and built MSR-VTT, a large-scale video to text dataset that is widely used worldwide. His works have led to many awards, including ACM SIGMM Outstanding Ph.D. Thesis Award 2015, ACM SIGMM Rising Star Award 2019, and IEEE TCMC Rising Star Award 2019. He is also an Associate Editor of IEEE Trans. on Multimedia.
\end{IEEEbiography}

\begin{IEEEbiography}[{\includegraphics[width=1in,height=1.25in,clip]{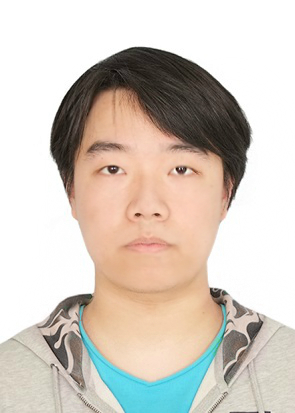}}]{Zhaofan Qiu}
	is currently a Researcher in Vision and Multimedia Lab at JD Explore Academy, Beijing, China. His research interests include large-scale video classification, semantic segmentation, and multimedia understanding. He has participated several large-scale video analysis competitions such as ActivityNet Large Scale Activity Recognition Challenge, and THUMOS Action Recognition Challenge. He was awarded the MSRA Fellowship in 2017. He received Ph.D. degree in 2020 from the University of Science and Technology of China (USTC), Hefei, China.
\end{IEEEbiography}

\begin{IEEEbiography}[{\includegraphics[width=1in,height=1.25in,clip]{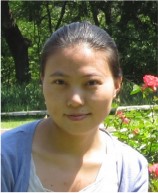}}]{Xinmei Tian}
	(M'13) is an Associate Professor in the CAS Key Laboratory of Technology in Geo-spatial Information Processing and Application System, University of Science and Technology of China. She received the B.E. degree and Ph.D. degree from the University of Science and Technology of China in 2005 and 2010, respectively. Her current research interests include multimedia information retrieval and machine learning. She received the Excellent Doctoral Dissertation of Chinese Academy of Sciences award in 2012 and the Nomination of National Excellent Doctoral Dissertation award in 2013.
\end{IEEEbiography}

\begin{IEEEbiography}[{\includegraphics[width=1in,height=1.2in,clip]{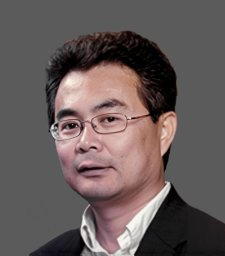}}]{Jiebo Luo}
	(S93, M96, SM99, F09) joined the Department of Computer Science at the University of Rochester in 2011, after a prolific career of over 15 years with Kodak Research. He has authored over 400 technical papers and holds over 90 U.S. patents. His research interests include computer vision, machine learning, data mining, social media,and biomedical informatics. He has served as the Program Chair of the ACM Multimedia 2010, IEEE CVPR 2012, ACM ICMR 2016, and IEEE ICIP 2017, and on the Editorial Boards of the IEEE TRANSACTIONS ON PATTERN ANALYSIS AND MACHINE INTELLIGENCE, IEEE TRANSACTIONS ON MULTIMEDIA, IEEE TRANSACTIONS ON CIRCUITS AND SYSTEMS FOR VIDEO TECHNOLOGY,  IEEE TRANSACTIONS ON BIG DATA,  Pattern Recognition, Machine Vision and Applications, and ACM Transactions on Intelligent Systems and Technology. He is also a Fellow of ACM, AAAI, SPIE and IAPR.
\end{IEEEbiography}

\begin{IEEEbiography}[{\includegraphics[width=1in,height=1.25in,clip,keepaspectratio]{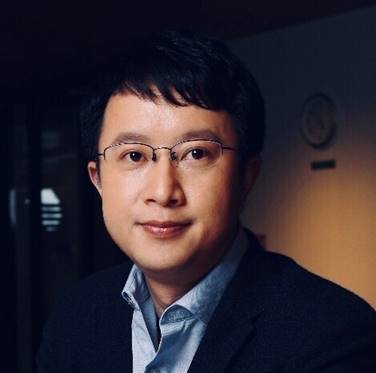}}]{Tao Mei}
	(M07-SM11-F19) is a Vice President with JD.COM and the Deputy Managing Director of JD Explore Academy, where he also serves as the Director of Computer Vision and Multimedia Lab. Prior to joining JD.COM in 2018, he was a Senior Research Manager with Microsoft Research Asia in Beijing, China. He has authored or co-authored over 200 publications (with 12 best paper awards) in journals and conferences, 10 book chapters, and edited five books. He holds over 25 US and international patents. He is or has been an Editorial Board Member of IEEE Trans. on Image Processing, IEEE Trans. on Circuits and Systems for Video Technology, IEEE Trans. on Multimedia, ACM Trans. on Multimedia Computing, Communications, and Applications, Pattern Recognition, etc. He is the General Co-chair of IEEE ICME 2019, the Program Co-chair of ACM Multimedia 2018, IEEE ICME 2015 and IEEE MMSP 2015.
	
	Tao received B.E. and Ph.D. degrees from the University of Science and Technology of China, Hefei, China, in 2001 and 2006, respectively. He is a Fellow of IEEE (2019), a Fellow of IAPR (2016), a Distinguished Scientist of ACM (2016), and a Distinguished Industry Speaker of IEEE Signal Processing Society (2017).
\end{IEEEbiography}

\end{document}